\documentclass[10pt,twocolumn,letterpaper]{article}

\makeatletter
\newcommand\notsotiny{\@setfontsize\notsotiny\@vipt\@viipt}
\makeatother

\usepackage[pagenumbers]{cvpr} 

\usepackage{graphicx}
\usepackage{amsmath}
\usepackage{amssymb}
\usepackage{booktabs}
\usepackage[table,xcdraw]{xcolor}
\usepackage{soul}
\usepackage[accsupp]{axessibility}  
\usepackage{titling}

\definecolor{myblue}{RGB}{33, 86, 131}
\definecolor{mypurple}{RGB}{112, 48, 160}
\definecolor{myred}{RGB}{252, 228, 214}
\definecolor{myyellow}{RGB}{255, 242, 204}

\usepackage[pagebackref,breaklinks,colorlinks]{hyperref}

\usepackage[capitalize]{cleveref}
\crefname{section}{Sec.}{Secs.}
\Crefname{section}{Section}{Sections}
\Crefname{table}{Table}{Tables}
\crefname{table}{Tab.}{Tabs.}
\graphicspath{{./images/}}

\def\cvprPaperID{2453} 
\def\confName{CVPR}
\def\confYear{2023}

\begin{document}

\title{NeRF-DS: Neural Radiance Fields for Dynamic Specular Objects}
\date{}

\author{
Zhiwen Yan \qquad Chen Li \qquad Gim Hee Lee\\
Department of Computer Science, National University of Singapore\\
{\tt\small \{yan.zhiwen, lichen\}@u.nus.edu} \qquad {\tt\small gimhee.lee@nus.edu.sg}
}
\maketitle

\begin{abstract}
   Dynamic Neural Radiance Field (NeRF) is a powerful algorithm capable of rendering photo-realistic novel view images from a monocular RGB video of a dynamic scene. Although it warps moving points across frames from the observation spaces to a common canonical space for rendering, dynamic NeRF does not model the change of the reflected color during the warping. As a result, this approach often fails drastically on challenging specular objects in motion. We address this limitation by reformulating the neural radiance field function to be conditioned on surface position and orientation in the observation space. This allows the specular surface at different poses to keep the different reflected colors when mapped to the common canonical space. Additionally, we add the mask of moving objects to guide the deformation field. As the specular surface changes color during motion, the mask mitigates the problem of failure to find temporal correspondences with only RGB supervision. We evaluate our model based on the novel view synthesis quality with a self-collected dataset of different moving specular objects in realistic environments.
   The experimental results demonstrate that our method significantly improves the reconstruction quality of moving specular objects from monocular RGB videos compared to the existing NeRF models. Our code and data are available at the project website \footnote{\url{https://github.com/JokerYan/NeRF-DS}}.
\end{abstract}

\section{Introduction}
\label{sec:intro}

Neural Radiance Fields (NeRF)\cite{mildenhall2020nerf} trained with multi-view images can synthesize novel views for 3D scenes with photo-realistic quality. NeRF predicts the volume density and view dependent color of the sampled spatial points in the scene with a multi-layer perceptron (MLP). Recent works 
such as Nerfies \cite{park2021nerfies} and NSFF \cite{li2020nsff} extend NeRF to reconstruct dynamic scenes from monocular videos. They resolve the lack of multi-view image supervision in dynamic scenes using a deformation field, which warps different observation spaces to a common canonical space.

\begin{figure}[t]
  \centering
   \includegraphics[width=0.8\linewidth]{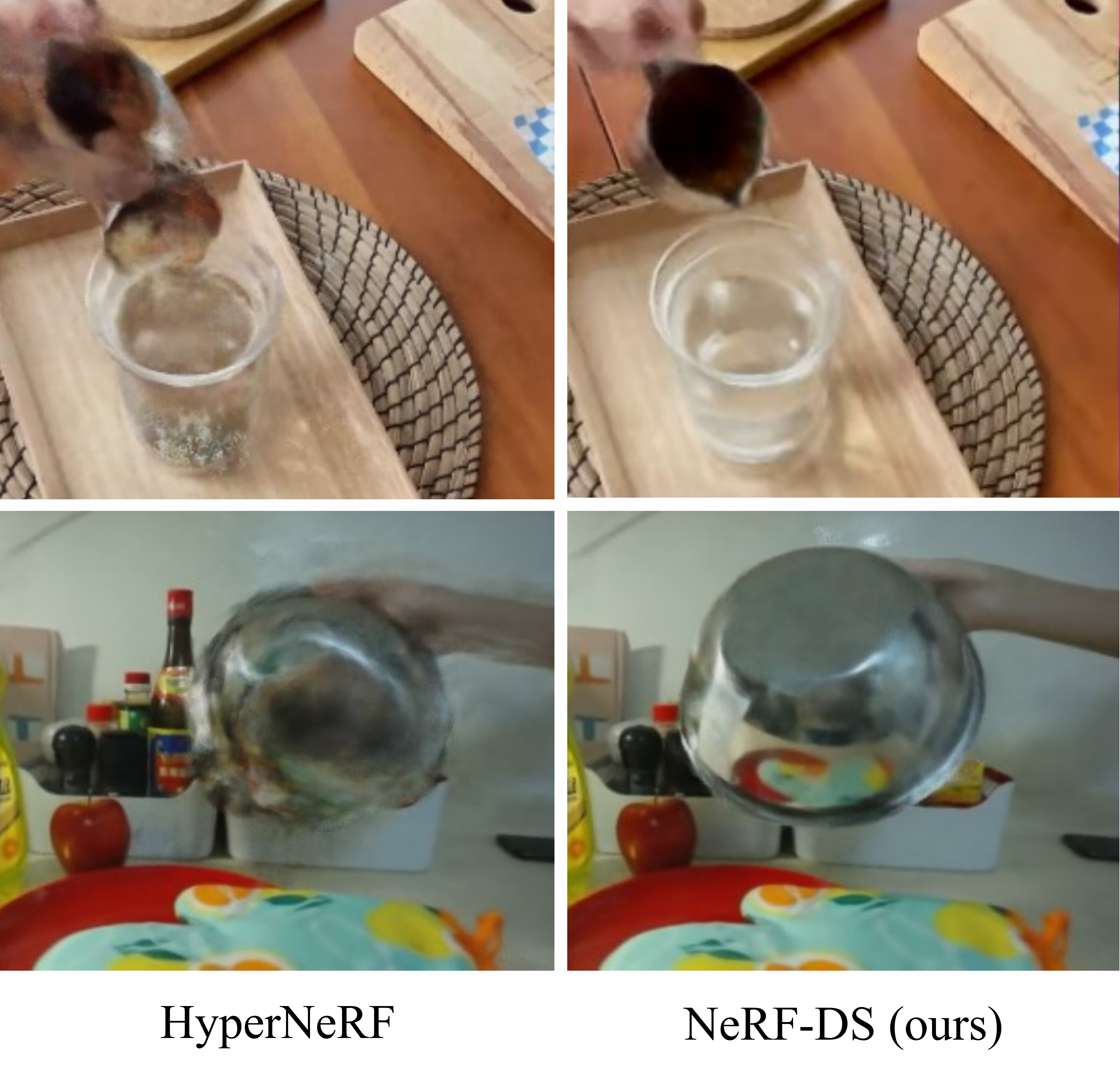}
   \caption[caption for teaser]{
   Comparison of novel views rendered by HyperNeRF~\cite{park2021hypernerf} (left) and our NeRF-DS (right), on the ``americano" scene in the HyperNeRF dataset\cite{park2021hypernerf}\footnotemark (top) and the ``basin" scene in our dynamic specular dataset (bottom).
   Our NeRF-DS model significantly improves the reconstruction quality by a surface-aware dynamic NeRF and a mask guided deformation field.} \vspace{-3mm}
   \label{fig:teaser}
\end{figure}

Despite showing promising results, we find that the existing dynamic NeRFs do not consider specular reflections during warping and often fail drastically on challenging dynamic specular objects as shown in ~\cref{fig:teaser}. The quality of dynamic specular object reconstruction is important because specular (e.g. metallic, plastic) surfaces are common in our daily environment 
and furthermore it indicates how accurate a dynamic NeRF represents the radiance field under motion or deformation. Previous works such as Ref-NeRF \cite{verbin2022refnerf} and NeRV \cite{nerv2021} have only focused on improving the specular reconstruction in static scenes. The problem of reconstructing dynamic specular objects with NeRF remain largely unexplored.

\footnotetext{The rendered frames come from the first 3 seconds of the ``americano" scene when the cup is rotating. This part of the video is not included in the HyperNeRF\cite{park2021hypernerf} qualitative results.}

We postulate that one of the reasons for dynamic models to fail on moving specular objects is because they do not consider the original surface information when rendering in a common canonical space. As suggested in rendering models such as Phong shading \cite{phong}, the specular color depends on the relative position and orientation of the surface with respect to the reflected environment. Nonetheless, existing dynamic NeRFs often ignore the original position and orientation of the surface when warping a specular object to a common canonical space for rendering. As the result, a point on a specular object reflecting different colors at different positions and orientations can cause conflicts when warped to a common canonical space.
Additionally, the key of existing dynamic models is to learn a deformation field for each frame such that correspondences can be established in a shared canonical space. However, the color of specular objects can vary significantly at different locations and orientations, which makes it hard to establish correspondences with the RGB supervision alone. These two limitations inevitably lead to the failure of existing dynamic models when applied to specular objects.

\begin{figure*}[t]
  \centering
   \includegraphics[width=1.0\textwidth]{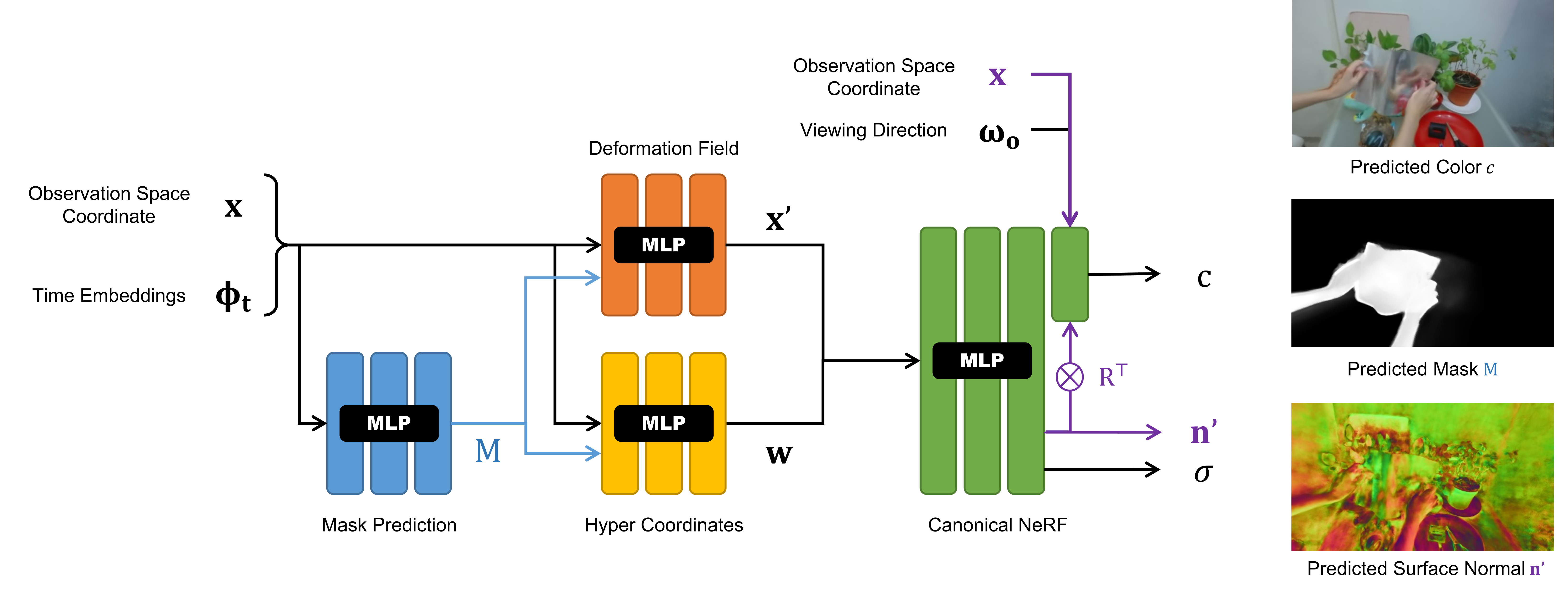}
   \caption{\textbf{An overview of our NeRF-DS.} We predict a 3D mask \textcolor{myblue}{$\operatorname{M}$} of the moving objects from observation space coordinate $\mathbf{x}$ and time embedding $\mathbf{\phi_t}$. Together with $\mathbf{x}$ and $\mathbf{\phi_t}$, the mask is used to guide the prediction of deformation field and hyper-coordinate (\textcolor{myblue}{blue arrows}). The canonical NeRF model takes in the canonical space coordinate $\mathbf{x'}$ the hyper-coordinate $\mathbf{w}$ to predict volume density $\sigma$ and canonical surface normal \textcolor{mypurple}{$\mathbf{n'}$}. The rotated surface normal \textcolor{mypurple}{$\mathbf{n}$} and coordinate \textcolor{mypurple}{$\mathbf{x}$} in observation space, together with the viewing direction $\mathbf{\omega}_o$ are fed to the color branch (\textcolor{mypurple}{purple arrows}) to predict color. Color and mask are supervised using the 2D ground truth after volumetric rendering, and surface normal is supervised by negative gradient of the volume density. }
   \label{fig:architecture}
\end{figure*}

In this paper, we introduce \textbf{NeRF-DS} (\cref{fig:architecture}) which models dynamic specular objects using a surface-aware dynamic NeRF and a mask guided deformation field to mitigate the two limitations mentioned above. 
1) Our NeRF-DS still warps the points from the observation space to a common canonical space and predicts their volume density. In contrast to other dynamic NeRFs, the color of each point is additionally conditioned on the spatial coordinate and surface normal in the \textit{observation space} before warping.
Corresponding points from different frames can share the same geometry, but reflect different colors determined by their original surface position and orientation.
2) Our NeRF-DS reuses the moving object mask from the camera registration stage as an additional input to the deformation field. This mask is a more consistent guidance for specular surfaces in motion compared to the constantly changing color. The mask is also a strong cue to the deformation field 
on the moving and static regions. 
As shown in~\cref{fig:teaser}, our proposed NeRF-DS reconstructs and renders dynamic specular scenes with significantly higher quality.

We implement our NeRF-DS on top of the \mbox{state-of-the-art} HyperNeRF \cite{park2021hypernerf} for dyanmic scenes. Since there are
very limited 
dynamic specular objects in the existing datasets, we collect another dynamic specular dataset for evaluation. 
Our dataset consists of a variety of moving/deforming specular objects in realistic environments. 
Experimental results on the dataset demonstrate that the NeRF-DS significantly improves the  quality of novel view rendering on dynamic specular objects. The images rendered by our NeRF-DS avoid many serious artifacts compared to the existing NeRF models. 

In summary, we have made the following contributions:
\begin{enumerate}
    \item A reparameterized dynamic NeRF that models dynamic specular surface with additional observation space coordinate and surface normal.
    \item A mask guided deformation field that improves deformation learned for dynamic specular objects.
    \item A dynamic specular scene dataset with training and testing monocular videos.
\end{enumerate}

\section{Related Work}
\paragraph{Neural Scene Representation and Rendering.}
The success of deep learning has led many works to explore suitable neural representations for 3D scene reconstruction and rendering. Explicit neural representations include point clouds\cite{fan2017point, yang2018foldingnet}, meshes\cite{bagautdinov2018modeling}, and voxels\cite{girdhar2016learning, tatarchenko2017octree}. Recent works have also explored various implicit neural representations of 3D scenes. Level set based representations map spatial coordinates to a signed distance function (SDF)\cite{Park_2019_CVPR, jiang2020local, yariv2020multiview} or occupancy fields \cite{mescheder2019occupancy}. These methods usually focus on the geometry reconstruction of the scene and requires additional neural representation of the texture \cite{niemeyer2020differentiable} to render the scene. 

An alternative implicit neural representation is the neural radiance field (NeRF) \cite{mildenhall2020nerf, barron2022mipnerf360, sitzmann2019srn}. NeRF directly represents the scene as a function that maps spatial coordinates and viewing angles to local point radiance. A differentiable volumetric rendering \cite{mildenhall2020nerf, kajiya1984ray} is performed to generate novel view images of the scene. NeRF can achieve photo-realistic novel view synthesis with only RGB supervision and known camera poses. Many extensions of NeRF are proposed, 
\eg acceleration\cite{wang2022fourier, reiser2021kilonerf}, scene scale\cite{tancik2022blocknerf, barron2022mipnerf360}, dynamic scenes\cite{tretschk2021nonrigid, park2021hypernerf, li2020nsff} and specular surface rendering\cite{verbin2022refnerf}. 

\vspace{-3mm}
\paragraph{Dynamic Scene Reconstruction.}
Dynamic scenes have objects moving in the foreground, objects undergoing deformation, or both. A simple reconstruction approach is to segment moving foreground and static background to reconstruct separately\cite{wong2021rigidfusion}. This method assumes the foreground is under rigid motion and cannot handle non-rigid deformation of the foreground object itself. A more general approach is to predict a canonical space and a temporal deformation field\cite{newcombe2015dynamicfusion, bozic2020nrtrack, lombardi2019neuralvolume, yu2017bodyfusion, niemeyer2019occupancyflow}. Many of these approaches require RGBD input \cite{newcombe2015dynamicfusion, yu2017bodyfusion} or multiple camera inputs \cite{lombardi2019neuralvolume} to resolve the ambiguity in reconstructing moving objects. 

Recent works \cite{park2021nerfies, park2021hypernerf, tretschk2021nonrigid, pumarola2020dnerf} based on the neural radiance field (NeRF) representation can jointly solve for canonical space and deformation field of dynamic scenes with only monocular RGB supervision. The canonical space in these works is usually a template NeRF as in static scenes, with the exception of HyperNeRF\cite{park2021hypernerf} which has additional hyper-coordinate input to model hyper canonical space. Another main difference among the existing dynamic NeRF models is the formulation of deformation field as a translation field \cite{tretschk2021nonrigid, pumarola2020dnerf, lombardi2019neuralvolume} or a special euclidean ($\operatorname{SE}(3)$) field \cite{park2021nerfies, park2021hypernerf}. Since NeRF is a coordinate based representation of the scene, the existing dynamic NeRFs mostly focus on warping spatial coordinates with the deformation field. They do not consider the changes to the object surface explicitly during the warping.

\vspace{-3mm}
\paragraph{Specular Surface Rendering.}
Rendering photorealistic images of specular or reflective surfaces is one of the most difficult problems in computer graphics. It usually requires the global illumination to be considered, traditionally achieved by expensive algorithms such as radiosity\cite{goral1984modeling,cohen1993radiosity}, ray tracing\cite{vlastimil2000rayshooting, purcell2002raytracing} or photon mapping\cite{jensen2001realistic}. To speed up the rendering, a technique called precomputed radiance transfer (PRT)\cite{sloan2002precomputed} is often used to precompute the lighting basis function in an environment map offline and rapidly sum them up during the online rendering phrase. For specular surfaces, 
the precomputation can be achieved by representing the reflection in spherical harmonics\cite{basri2003lambertian, ramamoorthi2001efficient, ramamoorthi2001signal}. 

In neural representations like NeRF, most works focusing on specular surface rendering follow the idea of precomputation. The reflection environment map can be considered ``precomputed" for each spatial point during the training. Some of the works based on the vanilla volumetric NeRF approximate the surface information needed for precomputation from volume density~\cite{verbin2022refnerf, boss2021neural, srinivasan2021nerv} or direct prediction~\cite{bi2020neural, zhang2021nerfactor, li2022neural}. Other works \cite{zhang2021ners, oechsle2021unisurf, yariv2020multiview} based on the signed distance function approximate the surface information from the signed distance. NeRFReN \cite{guo2022nerfren} splits the radiance transmitted and reflected components with a mask to render large flat reflective surfaces. Ref-NeRF\cite{verbin2022refnerf} proposes surface normal smoothing using MLP and directional encoding to further improve the performance. However, all the existing works in NeRF focusing on specular objects only consider static scenes instead of dynamic scenes. 

\vspace{5mm} 


\section{Dynamic NeRF Preliminaries}
NeRF\cite{mildenhall2020nerf} is a volumetric representation $F:(\mathbf{x}, \mathbf{\omega}_o)\rightarrow(\sigma, c)$\footnote{For simplicity in representations, we omit the $\sigma$ output of $F$ in the equations below unless otherwise specified.} of the scene. A multilayer-perceptron (MLP) is used to map the spatial position $\mathbf{x}$ to a volume density $\sigma(\mathbf{x})$ and bottleneck output $b(\mathbf{x})$. Another MLP head takes in bottleneck $b(\mathbf{x})$ and viewing direction (or outgoing radiance direction) $\mathbf{\omega}_o$ to predict the color $c(\mathbf{x}, \mathbf{\omega}_o)$ at the point:  
\begin{equation}
    c(\mathbf{x}, \mathbf{\omega}_o) = F(\mathbf{x}, \mathbf{\omega}_o)  .
\end{equation}

To render an image of the scene, $N$ samples $\mathbf{x_i}=\mathbf{o}-k_i\mathbf{\omega}_o$ are taken on each pixel ray $r$ from camera center $\mathbf{o}$. The color of the pixel $C(r)$ is the weighted sum of the colors at these sampled points, weighted by the product of accumulated transmittance $\alpha_i$ based on step size $\delta_i$ and local volume density along the ray:
\begin{subequations}
\begin{eqnarray}
    &\alpha_i = \exp(-\sum_{j=1}^{i-1} \sigma_i \delta_i),~w_i = \alpha_i(1-\exp(-\sigma_i\delta_i)),~~ \\
    &C(r)=\sum_{i=1}^N w_i \cdot c_i.
\end{eqnarray}
\end{subequations}

Dynamic NeRF \cite{park2021nerfies,park2021hypernerf,tretschk2021nonrigid,pumarola2020dnerf} reconstructs 3D dynamic scenes from monocular RGB camera footage. Since objects in a dynamic scene may be moving or deforming over time, only one frame is available for each moment of the scene. It is difficult to reconstruct the 3D structure of the scene without strict multi-view images. 
Consequently, most dynamic NeRFs transform the scene from an observation space at time $t$ to a common canonical space using a deformation field $T:\mathbf{x}\rightarrow \mathbf{x'}$.
Leveraging this common canonical space, images from different time and views can be used to reconstruct the scene
with a static NeRF model $F(\mathbf{x'},\mathbf{\omega}_o)$:
\begin{equation}
    c(\mathbf{x},\mathbf{\omega}_o,t)=F(T(\mathbf{x}, t), \mathbf{\omega}_o)=F(\mathbf{x'}, \mathbf{\omega}_o)  .
\end{equation}

In practice, the sampled observation space coordinate $\mathbf{x}$ and the time embedding $\mathbf{\phi}_t$ are fed into a deformation field prediction MLP to predict the canonical space coordinate $\mathbf{x'}$. HyperNeRF\cite{park2021hypernerf} additionally predicts a hyper canonical coordinate $\mathbf{w}$ from $\mathbf{x}$ and $\mathbf{\phi}_t$ using another MLP. The canonical coordinates $\mathbf{x'}$ and $\mathbf{w}$ are supplied to the canonical NeRF MLP to predict the volume density $\sigma$. A color prediction head of the canonical NeRF MLP takes in viewing direction $\mathbf{\omega}_o$ and outputs the color $c$. The existing dynamic NeRFs $F(\mathbf{x'}, \mathbf{\omega}_o)$ are under-parameterized when rendering dynamic specular objects. Particularly, the color should also depend on the observation space surface normal $\mathbf{n}$ and position $\mathbf{x}$. To this end, we propose to expand the model as $F(\mathbf{x'}, \mathbf{\omega}_o, \mathbf{x}, \mathbf{n})$. Refer to \cref{suraface_aware_dynamic_nerf} for more details.

\section{Our Method: NeRF-DS}
~\cref{fig:architecture} shows an illustration of our NeRF-DS which addresses the shortcomings of dynamic NeRFs for modeling the dynamic specular objects. Our NeRF-DS (on top of HyperNeRF\cite{park2021hypernerf}) includes a canonical NeRF conditioned on additional observation space position $\mathbf{x}$ and orientation $\mathbf{n}$ to predict the correct reflected color in the observation space (\cf \cref{suraface_aware_dynamic_nerf}).
$\mathbf{x}$ is obtained from ray samples and added with annealed positional encoding. $\mathbf{n}$ is obtained from warping the surface normal $\mathbf{n'}$ predicted in the canonical space. To better learn the correspondence and deformation field of specular surfaces, the deformation field and hyper coordinate prediction are guided with a mask $M$ of the moving objects (\cf \cref{mask_guided_deformation_field}).
$M$ is predicted by a mask prediction MLP and supervised by the 2D ground truth.

\subsection{Surface-Aware Dynamic NeRF}
\label{suraface_aware_dynamic_nerf}
In computer graphics, the rendering of specular surfaces is usually based on the rendering equation\cite{precomputation-rendering, kajiya1986rendering}:
\begin{equation}
    L_o(\mathbf{x},\mathbf{\omega}_o)=L_e(\mathbf{x}, \mathbf{\omega}_o)+\int_{\Omega}\rho(\mathbf{x}, \mathbf{\omega_i}, \mathbf{\omega}_o)L_i(\mathbf{x}, \mathbf{\omega_i})(\mathbf{\omega_i} \cdot \mathbf{n})d\mathbf{\omega_i}   ,
\end{equation}
where $L_o(\mathbf{x}, \mathbf{\omega}_o)$ is the outgoing radiance. The variables $\mathbf{x},~\mathbf{\omega_i},~\mathbf{\omega}_o$ and $\mathbf{n}$ represents the spatial coordinates, incident angle, outgoing angle, and surface normal, respectively. The first term $L_e(\mathbf{x}, \mathbf{\omega}_o)$ represents the emission light when the target object is a light source. The second term is a reflection component which integrates the outgoing reflected radiance of all incoming radiance $\mathbf{\omega_i}$ over the upper hemisphere $\Omega$ based on the BRDF $\rho$\cite{Nicodemus1977GeometricalCA} and the environment map $L_i$\cite{environment_map}.

In NeRF models, the color of radiance $L_o(\mathbf{x}, \mathbf{\omega}_o)$ is represented implicitly instead of integrated from all the reflected radiance explicitly. We can then simplify the reflection component to a function $L_r(\mathbf{x}, \mathbf{\omega}_o, \mathbf{n})$ and the rendering equation becomes:
\begin{equation}
    L_o(\mathbf{x},\mathbf{\omega}_o)=L_e(\mathbf{x}, \mathbf{\omega}_o)+L_r(\mathbf{x}, \mathbf{\omega}_o, \mathbf{n})   .
    \label{eq:emission_and_reflection}
\end{equation}
Under the assumption of no self-reflection, the reflected colors are all from the light source or objects in the static environment. The spatial coordinate $\mathbf{x}$, viewing direction $\mathbf{\omega}_o$ and surface normal $\mathbf{n}$ in \cref{eq:emission_and_reflection} are expressed in the observation space.

\begin{figure}[t]
  \centering
   \includegraphics[width=0.85\linewidth]{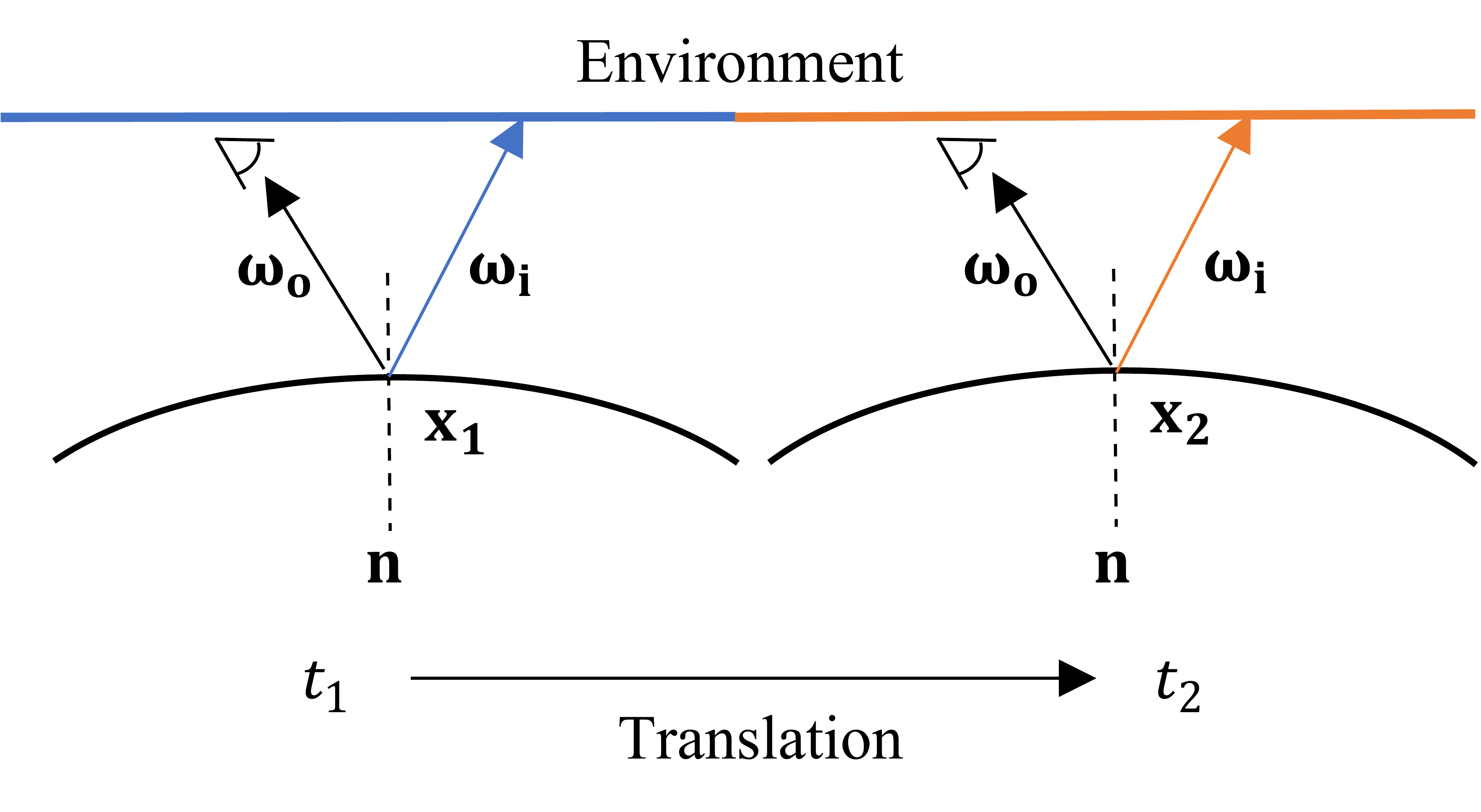}
   \caption{Existing dynamic NeRFs warp the translated points $\mathbf{x_1}$ and $\mathbf{x_2}$ to the same point in the canonical space, \ie $\mathrm{T}(\mathbf{x_1}, t_1)=\mathrm{T}(\mathbf{x_2}, t_2)=\mathbf{x'}$. As shown in the figure, the NeRF model $F(\mathbf{x'},\mathbf{\omega}_o)$ mistakenly renders them as the same color (assuming the same appearance code) instead of reflecting different colors.}
   \label{fig:specular_under_translation}
\end{figure}

\begin{figure}[t]
  \centering
   \includegraphics[width=0.6\linewidth]{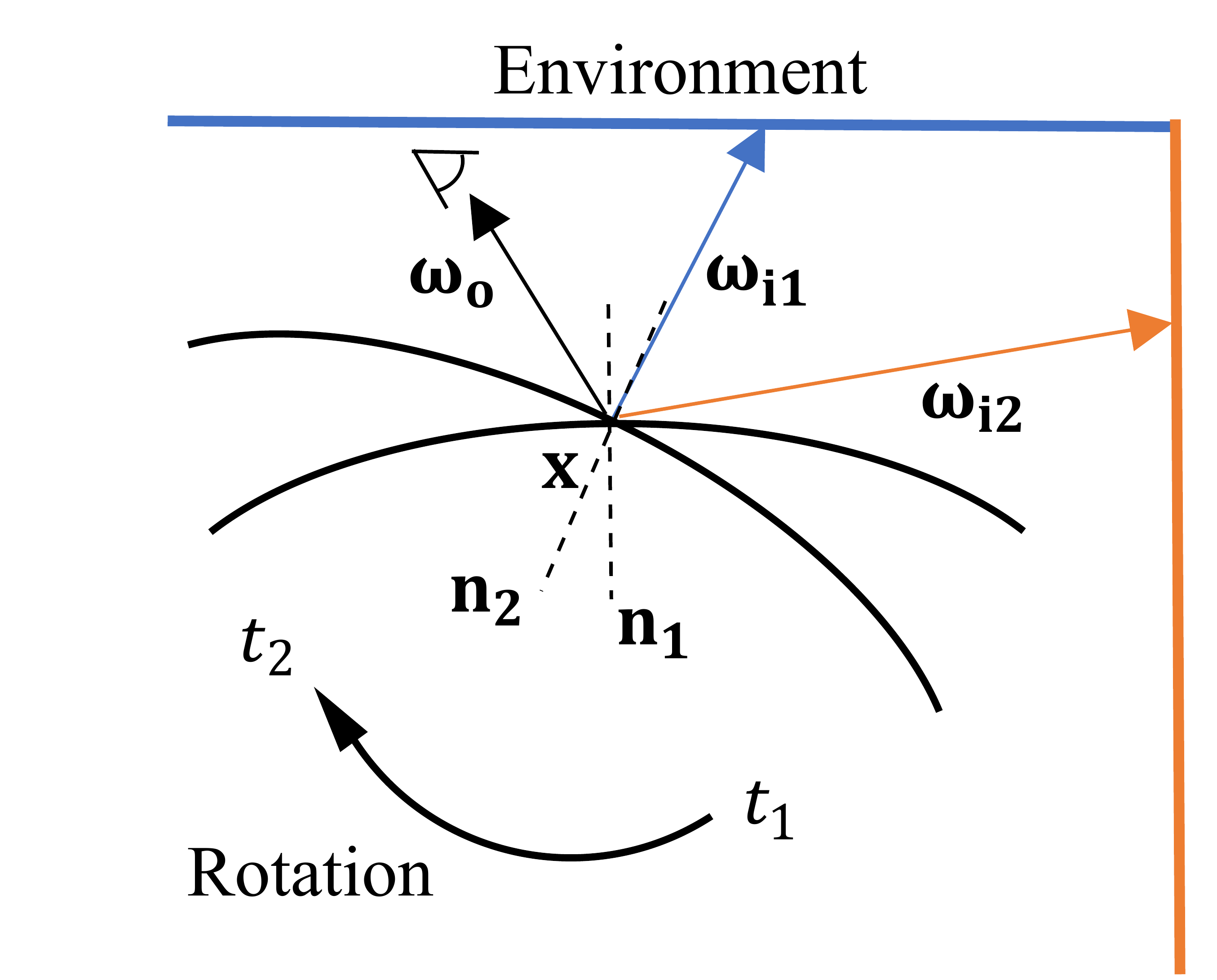}
   \caption{Existing dynamic NeRFs ignore the direction of surface normal. As shown in the figure, the NeRF model $F(\mathbf{x'},\mathbf{\omega}_o)$ mistakenly renders the point $\mathbf{x}$ before and after the rotation as the same color (assuming the same appearance code)
   instead of reflecting to different colors.} \vspace{-3mm}
   \label{fig:specular_under_rotation}
\end{figure}

In static scenes, the surfaces of the object do not move and therefore there is no difference between the observation and canonical spaces. As a result, the surface normal $\mathbf{n}$ can also be expressed as a function of $\mathbf{x}$ denoted by $N(\mathbf{x})$, and hence the rendering equation simplifies to:
\begin{equation}
    L_o(\mathbf{x},\mathbf{\omega}_o)=L_e(\mathbf{x}, \mathbf{\omega}_o)+L_r(\mathbf{x}, \mathbf{\omega}_o, N(\mathbf{x}))=F(\mathbf{x}, \mathbf{\omega}_o)   .
    \label{eq:static_render}
\end{equation}

In dynamic NeRFs, moving objects are first mapped from the observation spaces to a common canonical space to render. The points at the same canonical space position $\mathbf{x'}$ and viewing direction $\mathbf{\omega}_o$ are rendered the same color using NeRF MLP $F(\mathbf{x'}, \mathbf{\omega}_o)$. 
However, as described in the rendering equation from~\cref{eq:emission_and_reflection}, the color of the specular surface also depends on the observation space position $\mathbf{x}$ and surface normal $\mathbf{n}$. Points with the same $\mathbf{x'}$ and $\mathbf{\omega}_o$, but different $\mathbf{x}$ and $\mathbf{n}$ might 
reflect different colors. The existing dynamic NeRF in the form of $F:(\mathbf{x'}, \mathbf{\omega}_o) \rightarrow (\sigma, c)$ becomes an under-parameterized function in this case.
\cref{fig:specular_under_translation} and~\ref{fig:specular_under_rotation} illustrate two simple scenarios where the existing dynamic NeRF formulation fails on specular surfaces. 

We introduce a surface-aware dynamic NeRF following~\cref{eq:emission_and_reflection} to address the problem of under-parameterization in dynamic NeRFs. Surface information from the observation space is given to the canonical NeRF model to render the specular surface color. Specifically, we add the observation space coordinate $\mathbf{x}$ and surface normal $\mathbf{n}$ to the input of the NeRF color prediction branch (\textcolor{mypurple}{purple} in~\cref{fig:architecture}) while keeping the volume density prediction branch unchanged. The modified NeRF function can then be expressed as:
\begin{subequations}
\begin{align}
    L_o(\mathbf{x},\mathbf{\omega}_o)=L_e(\mathbf{x}, \mathbf{\omega}_o)&+L_r(\mathbf{x}, \mathbf{\omega}_o, \mathbf{n})=F(\mathbf{x'}, \mathbf{\omega}_o, \mathbf{x}, \mathbf{n}),~~ \\
    F:(\mathbf{x'},\mathbf{\omega}_o,\mathbf{x},\mathbf{n})& \rightarrow (\sigma(\mathbf{x'}),c(\mathbf{x'},\mathbf{\omega}_o,\mathbf{x},\mathbf{n})).
\end{align}
\end{subequations}

To prevent the model from directly rendering in the observation space and thus ignoring the shared canonical space, we follow \cite{park2021nerfies} to input the observation space coordinate $\mathbf{x}$ with annealed position encoding $\gamma_\tau(x)$:
\begin{subequations}
\begin{align}
    &z_j(\tau)=\frac{1-cos(\pi \cdot clamp(\tau-j,0,1))}{2}   , ~~\\
    \gamma_\tau(\mathbf{x})&=(\cdots,  z_k(\tau)sin(2^k\pi \mathbf{x}), z_k(\tau)cos(2^k\pi \mathbf{x}), \cdots)   .
\end{align}
\end{subequations}
The value of $\tau$ is initialized as 0 and slowly increased during training, so that $\mathbf{x}$ is completely cut off from the model in the early training stage. 

Unfortunately, the surface normal $\mathbf{n}$ cannot be directly extracted from volumetric models such as NeRF. To circumvent this problem, we first estimate the canonical space surface normal $\bar{\mathbf{n}}'$ with the negative gradient of volume density $\sigma$ with respect to the canonical space coordinate $\mathbf{x}$
\cite{nerd2021, nerv2021, verbin2022refnerf}:
\begin{equation}
    \bar{\mathbf{n}}'=-\frac{\nabla\sigma(\mathbf{x}')}{\Vert\nabla\sigma(\mathbf{x}')\Vert}.
\end{equation}
Nonetheless, the first order derivative of the volume density $\sigma$ is noisy without direct supervision. We thus use the estimated $\bar{\mathbf{n}}'$ to supervise a smoother predicted surface normal $\mathbf{n}'$ from the NeRF MLP and penalize any backward facing normal as in~\cite{verbin2022refnerf}, \ie:
\begin{subequations}
\begin{eqnarray}
    &\mathcal{L}_{norm}=\sum_{i}w_i\Vert \mathbf{n}'-\bar{\mathbf{n}}'\Vert^2   ,\\
    &\mathcal{L}_{backward}=\sum_{i}w_i \cdot \max(0, \mathbf{n}' \cdot -\mathbf{\omega}_o)   .
\end{eqnarray}
\end{subequations}

\begin{figure}[t]
  \centering
  \includegraphics[width=0.8\linewidth]{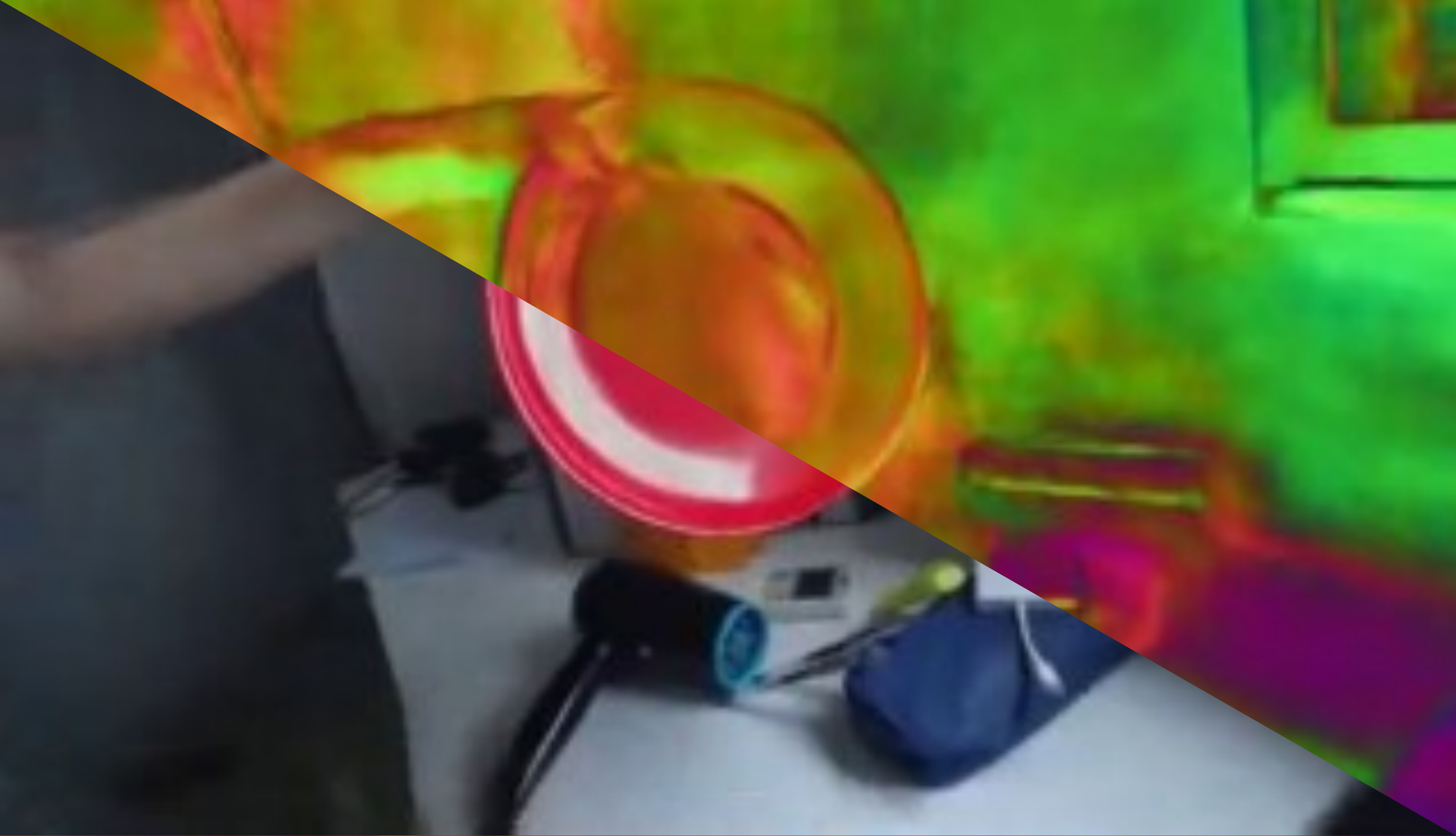}
   \caption{Surface normal in observation space, warped from the predicted canonical surface normal. The RGB values represent the $xyz$ components in the normalized surface norm vector. }
   \vspace{-3mm}
   \label{fig:norm_estimation}
\end{figure}

We use 3D special Euclidean group ($\operatorname{SE}(3)$) $\mathrm{T}(x)=[\mathrm{R} \mid \mathbf{t}] \mathbf{x}$ as our deformation field from the observation space to the canonical space. Finally, we can revert the canonical space surface normal $\mathbf{n}'$ back to observation space surface normal $\mathbf{n}$ using:
\begin{equation}
    \mathbf{n}=\mathrm{R}^\top \mathbf{n}'.
\end{equation}
Predicting and then warping the surface normal in canonical space ensures the surface normal consistency over time. The surface normals $\mathbf{n}_1$ and $\mathbf{n}_2$ of two corresponding points at time $t_1$ and $t_2$ are related by $\mathbf{n}_1=\mathrm{R}_1^\top \mathrm{R}_2 \mathbf{n}_2$. An example of final surface normal is illustrated in~\cref{fig:norm_estimation}.

\subsection{Mask Guided Deformation Field}
\label{mask_guided_deformation_field}
\begin{figure}[t]
  \centering
  \includegraphics[width=1\linewidth]{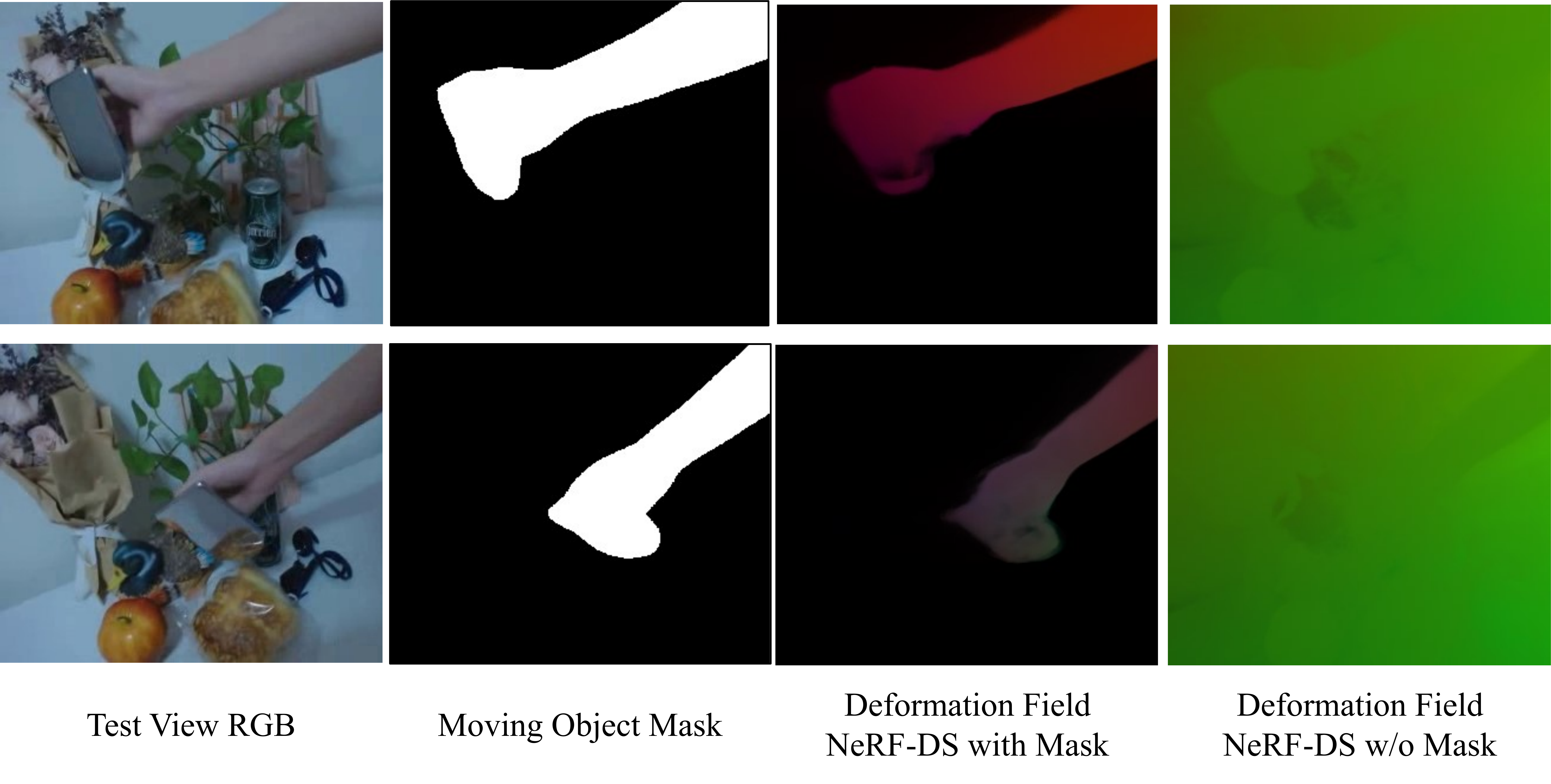}
   \caption{Comparison of deformation fields learned with and without mask as input, where the RGB values represent the $xyz$ components in the normalized deformation vector. The deformation field learned with masks well differentiates the moving foreground and the static background. The deformation field learned without masks fails to capture the foreground motion completely. }
   \vspace{-5mm}
   \label{fig:deformation_comparison}
\end{figure}

Most non-specular objects do not change color drastically when moving. However, the color of specular objects can change significantly at different positions and orientations as shown in~\cref{eq:emission_and_reflection}. The deformation of dynamic NeRFs is learned from RGB supervision only. Point correspondence can hardly be established if the color of the same point varies too much. As a result, the model often fails to learn the deformation field completely as shown in~\cref{fig:deformation_comparison}
 
To mitigate this issue, we introduce a mask guided deformation field using a 2D mask of the moving objects. Unlike the drastically changing color of specular surfaces, this mask remains consistent during the object motion. It provides meaningful guidance toward the deformation field prediction for specular surfaces. Additionally, the mask gives a strong cue to the deformation prediction network on the deforming regions.

We thus add a mask prediction network $G:\mathbf{x}\rightarrow \operatorname{M}$ that predicts the mask value at each 3D point in the observation space. The predicted mask $\operatorname{M}$ is fed to the deformation field and hyper-coordinate prediction networks (\textcolor{myblue}{blue} in~\cref{fig:architecture}). The predicted 3D mask is supervised by the 2D mask $\bar{\operatorname{M}}$ in training views using volumetric rendering:
\begin{equation}
    \mathcal{L}_{mask} =  \Vert (\sum_{i=1}^N w_i \cdot \operatorname{M}) - \bar{\operatorname{M}} \Vert^2.
\end{equation}
The mask prediction has more ambiguity than color prediction as the 2D mask is in binary values. We encourage the 3D mask to be predicted near the object surface by using sharper weights $w'_i$ instead of $w_i$. It is calculated by applying a Gaussian multiplier to weights $w_i$ for each sample $\mathbf{x}_i=\mathbf{o}+k_i\mathbf{\omega}_o$. The Gaussian $\mathcal{N}$ is centered at the maximum weights position $k_{\text{max}}$ and has a decreasing standard deviation $\beta$ during the training:
\begin{equation}
    w_i^* = w_i \cdot P(k_i|\mathcal{N}(k_{\text{max}},\beta)),~~  w_i'=w_i^*/(\sum_{j}w_j^*)  .
\end{equation}
As shown in~\cref{fig:deformation_comparison}, the mask guided deformation field results in a more meaningful deformation field predicted.

We note that this mask is already required by most dynamic NeRF during the camera pose registration \cite{park2021nerfies,park2021hypernerf,li2020nsff}. Moving foreground features must be masked out in structure-from-motion algorithms for correct registrations, thus we are not introducing additional input to the pipeline. Pose estimated without this mask can have significantly lower accuracy, especially when the moving part is large on the images. For example, camera poses estimated on our ``basin" scene without masks are 31.7\% deviated from the original poses after Procrustes alignment. HyperNeRF\cite{park2021hypernerf} trained on those poses performs 6.9\% worse in PSNR and 82.7\% worse in LPIPS.

\begin{figure*}[h]
  \centering
  \includegraphics[width=1\textwidth]{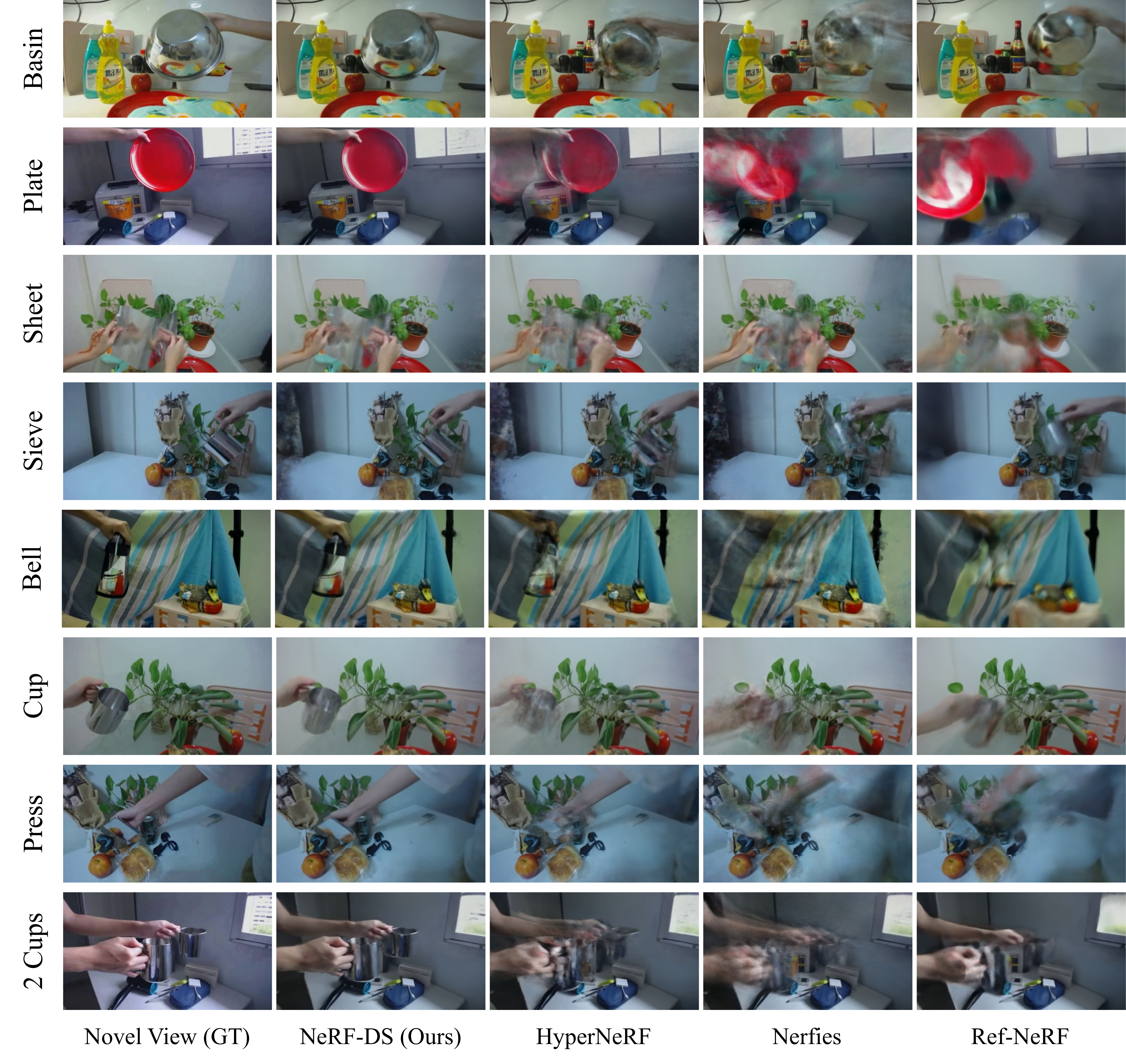}
   \vspace{-8mm}
   \caption{Qualitative comparisons between our NeRF-DS and the baselines on our dynamic specular dataset. Our NeRF-DS significantly reduces the severe tear-up and blurry artifacts compared to the baselines.}
   \vspace{-3mm}
   \label{fig:qualitative}
\end{figure*}


\section{Experiments}
\subsection{Dynamic Specular Dataset}
Existing dynamic NeRF datasets, \eg the scenes used in \cite{park2021nerfies, park2021hypernerf, li2020nsff, lombardi2019neuralvolume} include almost no moving specular objects. We thus collect a new dynamic specular dataset for evaluation. Our dataset consists of 8 scenes in everyday environments with various types of moving or deforming specular objects. Each scene contains two videos captured by two forward-facing cameras rigidly mounted together, similar to the setup in \cite{park2021nerfies}. The footage from one camera is used for training, and the other one is used for testing. This is different from the alternating training and testing cameras used in \cite{park2021nerfies}, which causes the ``unrealistic teleporting camera" problem mentioned in \cite{gao2022dynamic}. Each video contains $\sim500$ frames. 
The camera registration is performed using COLMAP~\cite{schoenberger2016mvs, schoenberger2016sfm} after applying a mask generated from MiVOS~\cite{cheng2021mivos}. This is the same mask used for mask prediction supervision in our mask guided deformation field module. See our supplementary for more details of the dataset. 

\subsection{Experimental Setups}
We evaluate the performance of our model based on the novel view synthesis quality on the dynamic specular dataset mentioned above. The video frames of one camera are used for training, and the model generates the novel view images at the pose of the other camera. The generated images are compared with the ground truth test view images to calculate the following quantitative metrics: MS-SSIM\cite{wang2003multiscale}, PSNR, LPIPS\cite{zhang2018unreasonable} as in previous works\cite{park2021hypernerf, nerv2021, li2020nsff}. The average score across all the frames are reported. 

We compare our model with the baseline models of HyperNeRF\cite{park2021hypernerf}, Nerfies\cite{park2021nerfies} and Ref-NeRF\cite{verbin2022refnerf}. HyperNeRF achieves the \mbox{state-of-the-art} dynamic NeRF performance by introducing hyper-coordinates. Nerfies is a representative dynamic NeRF model with a standard canonical + deformation setup and outperforms many other models with similar designs. Ref-NeRF achieves the \mbox{state-of-the-art} reconstruction quality of static specular surfaces. These three baseline models well represent the SOTA performance of dynamic and specular scene reconstruction with NeRF.

Our implementation of NeRF-DS is based on HyperNeRF. The new mask prediction network is a 6-layer MLP with a width of 64. The final output of the mask prediction undergoes a ReLU activation. All baseline models and our models are trained following the respective official configuration for 250k iterations. All training and rendering are performed in $480 \times 270$ resolution. More implementation details and experiment setup are in our supplementary. 

\subsection{Evaluation Results}
In this section, we present the quantitative and qualitative results of our method compared to the baselines and the ablation models. Several videos of the rendered sequences are included in the supplementary.

\begin{table*}[]
\centering
\footnotesize
\begin{tabular}{l|m{0.5cm}m{0.5cm}m{0.5cm}m{0.5cm}m{0.5cm}m{0.5cm}m{0.5cm}m{0.5cm}m{0.5cm}m{0.5cm}m{0.5cm}m{0.5cm}m{0.5cm}m{0.5cm}m{0.5cm}}
\hline
                    & \multicolumn{6}{c|}{Mean}                                                                                                                                                                                                             & \multicolumn{3}{c|}{Bell}                                                                                         & \multicolumn{3}{c|}{Plate}                                                                                        & \multicolumn{3}{c}{Sheet}                                                                    \\
                    & \multicolumn{2}{c}{\notsotiny{MS-SSIM$\uparrow$}}                                  & \multicolumn{2}{c}{\notsotiny{PSNR$\uparrow$}}                                                           & \multicolumn{2}{c|}{\notsotiny{LPIPS$\downarrow$}}                                                        & \notsotiny{MS-SSIM$\uparrow$}                       & \notsotiny{PSNR$\uparrow$}                         & \multicolumn{1}{m{0.5cm}|}{\notsotiny{LPIPS$\downarrow$}}                         & \notsotiny{MS-SSIM$\uparrow$}                       & \notsotiny{PSNR$\uparrow$}                         & \multicolumn{1}{m{0.5cm}|}{\notsotiny{LPIPS$\downarrow$}}                         & \notsotiny{MS-SSIM$\uparrow$}                       & \notsotiny{PSNR$\uparrow$}                         & \notsotiny{LPIPS$\downarrow$}                         \\ \hline
Ref-NeRF \cite{verbin2022refnerf}            & \multicolumn{2}{c}{.640}                                    & \multicolumn{2}{c}{19.2}                                                           & \multicolumn{2}{c|}{.354}                                                        & .564                         & 18.5                         & \multicolumn{1}{m{0.5cm}|}{.420}                         & .513                         & 15.3                         & \multicolumn{1}{m{0.5cm}|}{.464}                         & .673                         & 21.1                         & .296                         \\
Nerfies \cite{park2021nerfies}             & \multicolumn{2}{c}{.689}                                    & \multicolumn{2}{c}{19.7}                                                           & \multicolumn{2}{c|}{.381}                                                        & .696                         & 19.9                         & \multicolumn{1}{m{0.5cm}|}{.389}                         & .489                         & 15.4                         & \multicolumn{1}{m{0.5cm}|}{.599}                         & .834                         & 23.6                         & .183                         \\
HyperNeRF \cite{park2021hypernerf}           & \multicolumn{2}{c}{.849}                                    & \multicolumn{2}{c}{22.7}                                                           & \multicolumn{2}{c|}{.192}                                                        & \cellcolor[HTML]{FCE4D6}.884 & \cellcolor[HTML]{FCE4D6}24.0 & \multicolumn{1}{m{0.5cm}|}{.159}                         & .714                         & 18.1                         & \multicolumn{1}{m{0.5cm}|}{.359}                         & .874                         & 24.3                         & .148                         \\ \hline
NeRF-DS (Ours)      & \multicolumn{2}{c}{\cellcolor[HTML]{FCE4D6}.890}            & \multicolumn{2}{c}{\cellcolor[HTML]{FCE4D6}23.4}                                   & \multicolumn{2}{c|}{\cellcolor[HTML]{FCE4D6}.135}                                & .872                         & 23.3                         & \multicolumn{1}{m{0.5cm}|}{\cellcolor[HTML]{FCE4D6}.134} & \cellcolor[HTML]{FCE4D6}.867 & \cellcolor[HTML]{FCE4D6}20.8 & \multicolumn{1}{m{0.5cm}|}{\cellcolor[HTML]{FCE4D6}.164} & \cellcolor[HTML]{FCE4D6}.918 & \cellcolor[HTML]{FCE4D6}25.7 & \cellcolor[HTML]{FCE4D6}.115 \\
NeRF-DS w/o Surface & \multicolumn{2}{c}{\cellcolor[HTML]{FFF2CC}.881}            & \multicolumn{2}{c}{\cellcolor[HTML]{FFF2CC}23.3}                                   & \multicolumn{2}{c|}{\cellcolor[HTML]{FFF2CC}.142}                                & .867                         & 23.2                         & \multicolumn{1}{m{0.5cm}|}{\cellcolor[HTML]{FFF2CC}.141} & \cellcolor[HTML]{FFF2CC}.861 & \cellcolor[HTML]{FCE4D6}20.8 & \multicolumn{1}{m{0.5cm}|}{\cellcolor[HTML]{FFF2CC}.171} & \cellcolor[HTML]{FFF2CC}.905 & \cellcolor[HTML]{FFF2CC}25.2 & \cellcolor[HTML]{FFF2CC}.124 \\
NeRF-DS w/o Mask    & \multicolumn{2}{c}{\cellcolor[HTML]{FFF2CC}.881}            & \multicolumn{2}{c}{\cellcolor[HTML]{FFF2CC}23.3}                                   & \multicolumn{2}{c|}{.153}                                                        & \cellcolor[HTML]{FFF2CC}.887 & \cellcolor[HTML]{FFF2CC}23.9 & \multicolumn{1}{m{0.5cm}|}{.138}                         & .855                         & 20.5                         & \multicolumn{1}{m{0.5cm}|}{.196}                         & .887                         & 24.7                         & .141                         \\ \hline
                    &                               &                              &                                                    &                               &                              &                                                    &                               &                              &                                                    &                               &                              &                                                    &                               &                              &                               \\ \hline
                    & \multicolumn{3}{c|}{Sieve}                                                                                        & \multicolumn{3}{c|}{Basin}                                                                                        & \multicolumn{3}{c|}{Cup}                                                                                          & \multicolumn{3}{c|}{Press}                                                                                        & \multicolumn{3}{c}{2 Cups}                                                                   \\
                    & \notsotiny{MS-SSIM$\uparrow$}                       & \notsotiny{PSNR$\uparrow$}                         & \multicolumn{1}{m{0.5cm}|}{\notsotiny{LPIPS$\downarrow$}}                         & \notsotiny{MS-SSIM$\uparrow$}                       & \notsotiny{PSNR$\uparrow$}                         & \multicolumn{1}{m{0.5cm}|}{\notsotiny{LPIPS$\downarrow$}}                         & \notsotiny{MS-SSIM$\uparrow$}                       & \notsotiny{PSNR$\uparrow$}                         & \multicolumn{1}{m{0.5cm}|}{\notsotiny{LPIPS$\downarrow$}}                         & \notsotiny{MS-SSIM$\uparrow$}                       & \notsotiny{PSNR$\uparrow$}                         & \multicolumn{1}{m{0.5cm}|}{\notsotiny{LPIPS$\downarrow$}}                         & \notsotiny{MS-SSIM$\uparrow$}                       & \notsotiny{PSNR$\uparrow$}                         & \notsotiny{LPIPS$\downarrow$}                         \\ \hline
Ref-NeRF \cite{verbin2022refnerf}            & .815                         & 22.1                         & \multicolumn{1}{m{0.5cm}|}{.220}                         & .643                         & 18.0                         & \multicolumn{1}{m{0.5cm}|}{.319}                         & .705                         & 20.5                         & \multicolumn{1}{m{0.5cm}|}{.318}                         & .679                         & 21.3                         & \multicolumn{1}{m{0.5cm}|}{.341}                         & .527                         & 16.5                         & .454                         \\
Nerfies \cite{park2021nerfies}             & .823                         & 21.8                         & \multicolumn{1}{m{0.5cm}|}{.232}                         & .635                         & 18.1                         & \multicolumn{1}{m{0.5cm}|}{.368}                         & .750                         & 20.7                         & \multicolumn{1}{m{0.5cm}|}{.293}                         & .720                         & 21.3                         & \multicolumn{1}{m{0.5cm}|}{.377}                         & .563                         & 17.1                         & .605                         \\
HyperNeRF \cite{park2021hypernerf}           & .909                         & 25.0                         & \multicolumn{1}{m{0.5cm}|}{.129}                         & .829                         & 20.2                         & \multicolumn{1}{m{0.5cm}|}{.168}                         & .896                         & 24.1                         & \multicolumn{1}{m{0.5cm}|}{.138}                         & .873                         & 25.4                         & \multicolumn{1}{m{0.5cm}|}{.164}                         & .809                         & 20.1                         & .272                         \\ \hline
NeRF-DS (Ours)      & \cellcolor[HTML]{FCE4D6}.935 & \cellcolor[HTML]{FFF2CC}26.1 & \multicolumn{1}{m{0.5cm}|}{\cellcolor[HTML]{FFF2CC}.108} & \cellcolor[HTML]{FCE4D6}.868 & \cellcolor[HTML]{FCE4D6}20.3 & \multicolumn{1}{m{0.5cm}|}{\cellcolor[HTML]{FCE4D6}.127} & \cellcolor[HTML]{FFF2CC}.916 & \cellcolor[HTML]{FFF2CC}24.5 & \multicolumn{1}{m{0.5cm}|}{\cellcolor[HTML]{FFF2CC}.118} & \cellcolor[HTML]{FCE4D6}.911 & \cellcolor[HTML]{FCE4D6}26.4 & \multicolumn{1}{m{0.5cm}|}{\cellcolor[HTML]{FCE4D6}.123} & \cellcolor[HTML]{FFF2CC}.836 & \cellcolor[HTML]{FFF2CC}20.4 & \cellcolor[HTML]{FCE4D6}.193 \\
NeRF-DS w/o Surface & \cellcolor[HTML]{FCE4D6}.935 & \cellcolor[HTML]{FCE4D6}26.2 & \multicolumn{1}{m{0.5cm}|}{\cellcolor[HTML]{FCE4D6}.107} & \cellcolor[HTML]{FCE4D6}.868 & \cellcolor[HTML]{FCE4D6}20.3 & \multicolumn{1}{m{0.5cm}|}{\cellcolor[HTML]{FFF2CC}.128} & \cellcolor[HTML]{FCE4D6}.918 & \cellcolor[HTML]{FCE4D6}24.6 & \multicolumn{1}{m{0.5cm}|}{\cellcolor[HTML]{FCE4D6}.117} & .886                         & 25.8                         & \multicolumn{1}{m{0.5cm}|}{\cellcolor[HTML]{FFF2CC}.140} & .810                         & 20.3                         & \cellcolor[HTML]{FFF2CC}.211 \\
NeRF-DS w/o Mask    & .928                         & 26.0                         & \multicolumn{1}{m{0.5cm}|}{.112}                         & .835                         & 20.1                         & \multicolumn{1}{m{0.5cm}|}{.149}                         & .912                         & 24.4                         & \multicolumn{1}{m{0.5cm}|}{.122}                         & \cellcolor[HTML]{FFF2CC}.894 & \cellcolor[HTML]{FFF2CC}26.1 & \multicolumn{1}{m{0.5cm}|}{.142}                         & \cellcolor[HTML]{FCE4D6}.849 & \cellcolor[HTML]{FCE4D6}20.9 & .220                         \\ \hline
\end{tabular}
\caption{Quantitative comparisons between our NeRF-DS against the baselines and ablations of our model. The \colorbox{myred}{best} and \colorbox{myyellow}{second best} results for each scene are color coded. ``NeRF-DS w/o Surface" is our model without the surface-aware dynamic NeRF module. ``NeRF-DS w/o Mask" is our model without the mask guided defromation field module.}
\vspace{-3mm}
\label{tab:quantitative_results}
\end{table*}

\vspace{-3mm}
\paragraph{Qualitative Results.}
\label{qualitative_results}
We present the qualitative results of novel view synthesis in~\cref{fig:qualitative}. The HyperNeRF and Nerfies models tend to reconstruct dynamic specular objects with severe geometric artifacts. The rendered objects are blurry or torn apart along the moving trajectory. This can be attributed to two reasons: 1) The model struggles to capture the specular color without any observation space surface normal and location information. 2) The color of the specular object at the same point varies a lot, which makes it hard for the existing dynamic NeRF to learn a meaningful transformation field. This causes the sample points to warp to the wrong locations and resulting in the ``torn-up" effect. 
Ref-NeRF also produces very blurry or torn-up results on dynamic specular objects. This is because Ref-NeRF assumes the scene to be static for all video frames. Since the object is actually moving, direct triangulation without warping would fail and thus leading to wrong prediction of the object geometry. 
Our NeRF-DS renders dynamic specular scenes with much fewer geometric artifacts. The reflected color on the specular surfaces is also relatively accurate. With the surface aware dynamic NeRF, the same canonical position is allowed to be mapped to different observation space positions reflecting different colors. The mask of moving objects guides the points in the observation space to learn the correct deformation mapping to the canonical space.  As a result, the scene reconstructed by NeRF-DS is free from the ``torn-up" effect present in other dynamic NeRFs. The reflected color is also more accurately controlled by the position and orientation of the surface.

\vspace{-3mm}
\paragraph{Quantitative Results.}
\label{quantiative_results}
We report the quantitative results in~\cref{tab:quantitative_results}. 
Note that LPIPS is taken to be a better measure of the construction quality compared to MS-SSIM and PSNR in previous works~\cite{park2021nerfies, park2021hypernerf}. During our qualitative evaluation, we also observe that MS-SSIM and PSNR are sometimes not affected significantly by blurry predictions. As shown in~\cref{tab:quantitative_results}, our NeRF-DS outperforms all baseline models by a significant margin evaluated with LPIPS. NeRF-DS also has better MS-SSIM and PSNR scores in most scenes and the overall average. 

\vspace{-3mm}
\paragraph{Ablation Study.}
We evaluate the contributions of the two proposed components of our model: the surface-aware dynamic NeRF, and the mask guided deformation field by removing each of them at a time. The models without surface information and without masks are denoted as ``NeRF-DS w/o Surface" and ``NeRF-DS w/o Mask", respectively. We report the quantitative comparisons in~\cref{tab:quantitative_results}, and the qualitative comparisons in the supplementary. The results suggest that the performance drops when either component is removed, which verify the contribution of each component. Additionally, the superior performance of both ablation models compared to baselines further supports the effectiveness of our proposed methods.

\section{Limitations}
Although NeRF-DS significantly improves the reconstruction quality of dynamic specular objects, it relies on accurate surface normal predictions. Unfortunately, we have observed that the geometry of the surface predicted by NeRF can be misled by the reflected texture. The predicted surface normal takes the shape of the reflected textures instead of the surface geometry. This problem is more severe in dynamic specular scenes than in static specular scenes due to the lack of strict geometry constraints. We leave the exploration of surface priors or more constrained deformation models to our future work.

\section{Conclusion}
Our proposed NeRF-DS extends the prior dynamic NeRF to reconstruct and render dynamic specular scenes more accurately. We introduce surface-aware dynamic NeRF to address the under-parameterization problem of rendering specular surfaces in the canonical space. We further design a mask guided deformation field to learn better correspondence under constant color changes. Both components are essential to model reflected colors during the warping to the canonical space. Our NeRF-DS achieves a better novel view synthesis quality compared to the prior dynamic and reflective NeRFs on dynamic specular scenes. 

\paragraph{Acknowledgement.}
This research is supported by the National Research Foundation, Singapore under its AI Singapore Programme (AISG Award
No: AISG2-RP-2021-024), and the Tier 2 grant MOE-T2EP20120-0011 from the
Singapore Ministry of Education.

{\small
\bibliographystyle{ieee_fullname}
\bibliography{egbib}
}

\clearpage
\setcounter{section}{0}
\title{Supplementary for  \\ NeRF-DS: Neural Radiance Fields for Dynamic Specular Objects}
\date{}
\author{}
\maketitle

\section{Qualitative Result Videos}
We include a few videos rendered by our model and the baseline models in the supplementary zip file as a better demonstration of the qualitative performance comparison.

\section{Implementation Details}
\begin{figure}[th]
  \centering
   \includegraphics[width=1.0\linewidth]{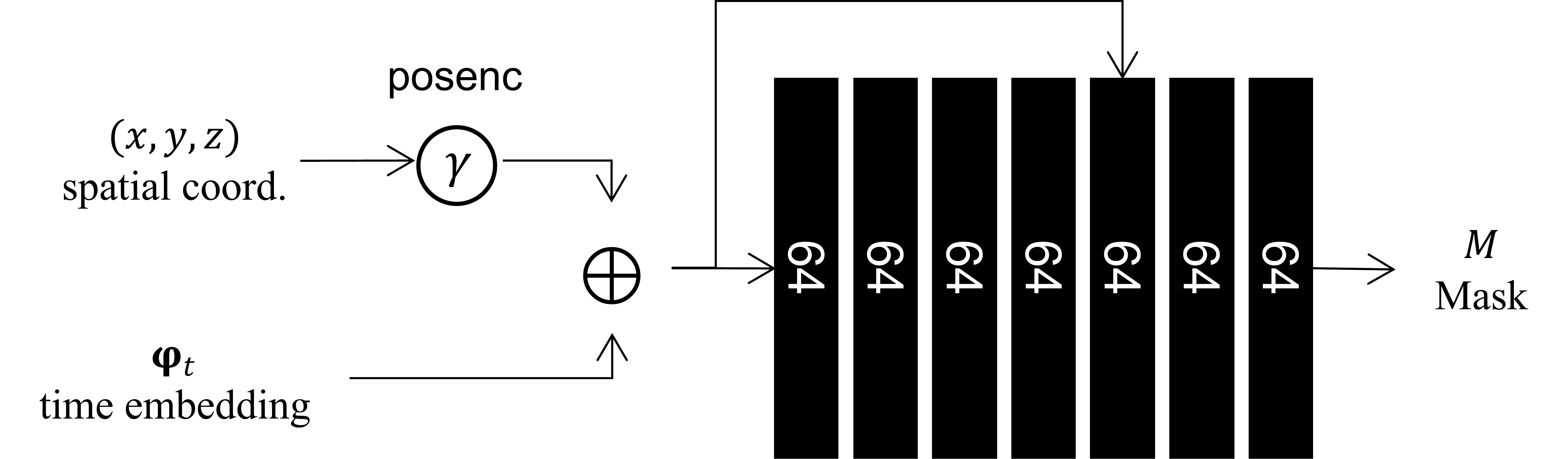}
   \caption{Architecture of the mask prediction module.}
   \label{fig:module_mask}
\end{figure}

\begin{figure}[th]
  \centering
   \includegraphics[width=1.0\linewidth]{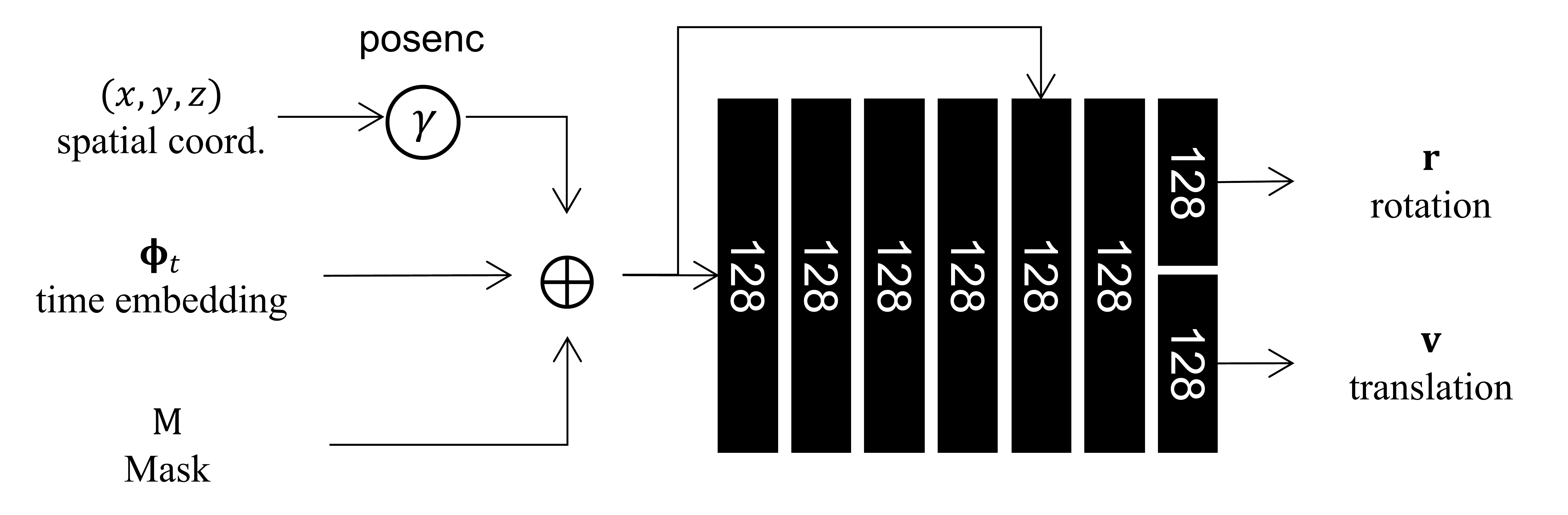}
   \caption{Architecture of the deformation field prediction module.}
   \label{fig:module_deformation}
\end{figure}

\begin{figure}[th]
  \centering
   \includegraphics[width=1.0\linewidth]{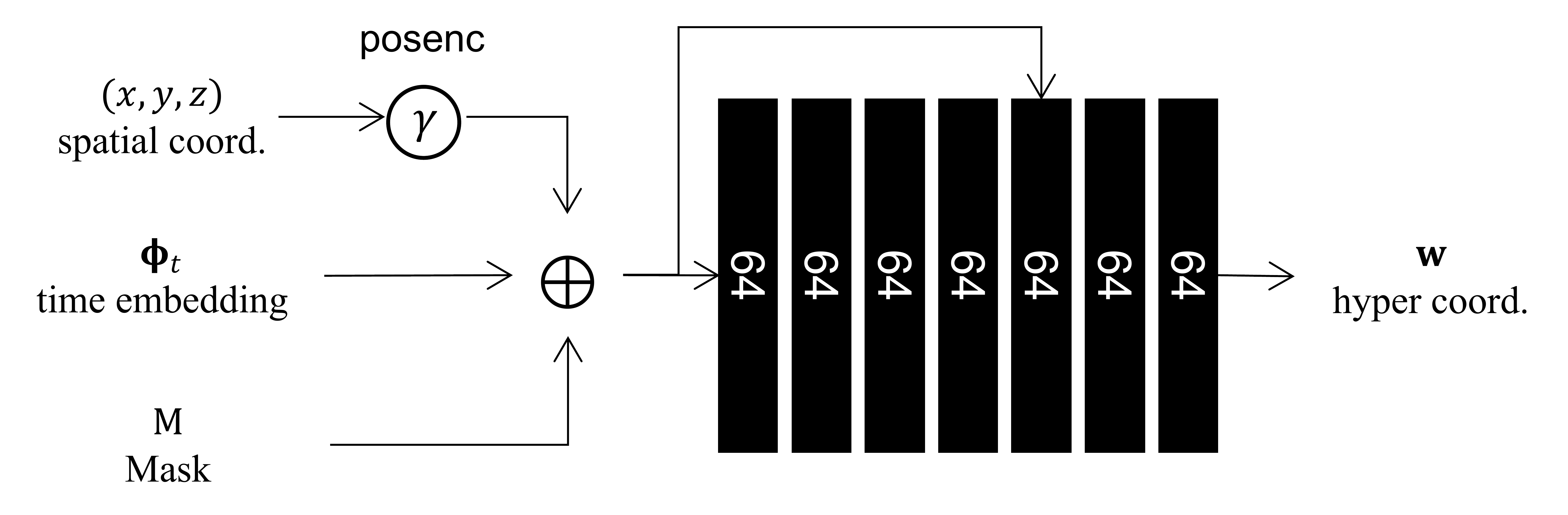}
   \caption{Architecture of the hyper coordinates prediction module.}
   \label{fig:module_hyper}
\end{figure}

\begin{figure}[th]
  \centering
   \includegraphics[width=1.0\linewidth]{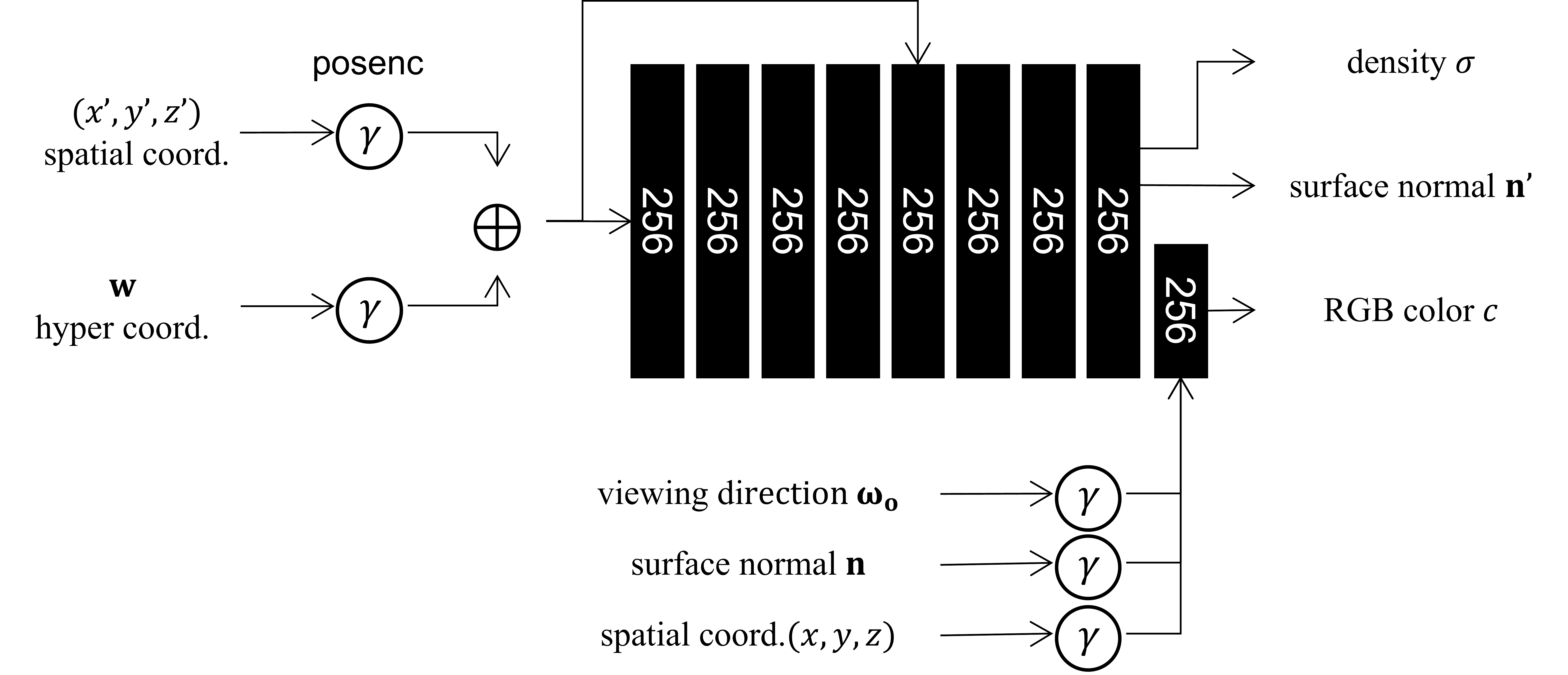}
   \caption{Architecture of the canonical NeRF module.}
   \label{fig:module_nerf}
\end{figure}

The details of the mask prediction module~(\cref{fig:module_mask}), deformation prediction module~(\cref{fig:module_deformation}), hyper coordinate prediction module~(\cref{fig:module_hyper}) and canonical NeRF~(\cref{fig:module_nerf}) module are illustrated in the respective figure.
Positional encoding is performed on spatial coordinates $\mathbf{x}$, $\mathbf{x'}$, viewing direction $\mathbf{\omega_o}$ and surface normal $\mathbf{n}$. Different encoding widths and annealing widths are used for different input as shown in~\cref{tab:posenc}.
The Gaussian applied to the weights $w'$ for mask volumetric rendering has an exponentially decreasing standard deviation $\beta$ from 1 to 0.1 during the first 30k iterations. Then it stays constant at 0.1 for the rest of the training. 

\subsection{Details of Ref-NeRF Experiments}
We use the official integrated Ref-NeRF\cite{park2021nerfies} code from Multi-NeRF\cite{multinerf2022}. To accommodate our dynamic specular dataset, we slightly adjust the scene offset and scaling logic to ensure the scene is well centered and bounded. 

\subsection{Parameter and Training Time}
The full model contains 1.45M parameters, compared to the 1.30M parameters of the baseline model. The experiment with 480x270 resolution videos takes 6 hours to train on 4 RTX A5000 GPUs, compared to the 5 hours training time of the baseline model. 

\begin{table}[]
\scriptsize
\begin{tabular}{|l|ccccc|}
\hline
                                    & width & anneal & delay iter. & inc. iter. & inc. func. \\ \hline
$\mathbf{x}$ to mask                & 4     & Yes    & 0k          & 50k        & linear     \\
$\mathbf{x}$ to deformation         & 4     & Yes    & 0k          & 50k        & linear     \\
$\mathbf{x}$ to hyper coord.        & 6     & No     & N/A         & N/A        & N/A        \\
$\mathbf{x}$ to color branch        & 4     & Yes    & 50k         & 50k        & linear     \\
$\mathbf{x'}$ to NeRF               & 8     & No     & N/A         & N/A        & N/A        \\
$\mathbf{w}$ to NeRF                & 1     & No     & N/A         & N/A        & N/A        \\
$\mathbf{\omega_o}$ to color branch & 4     & No     & N/A         & N/A        & N/A        \\
$\mathbf{n}$ to color branch        & 4     & Yes    & 10k         & 2k         & linear     \\ \hline
\end{tabular}
\caption{Details of the positional encoding and annealing of each input. ``Width" indicates the highest $k$ in $sin(2^k\pi\mathbf{x})$ sequence. ``Anneal" indicates whether annealing coefficient $z_j(\tau)$ for positional encoding is used. If annealing is used, ``delay iter." is the number of iterations where $\tau$ stays 0 at the start of the training. ``inc. iter." and ``inc. func." are the number of increasing iterations and function after the delay.}
\label{tab:posenc}
\end{table}

\begin{figure*}[h]
  \centering
   \includegraphics[width=1.0\textwidth]{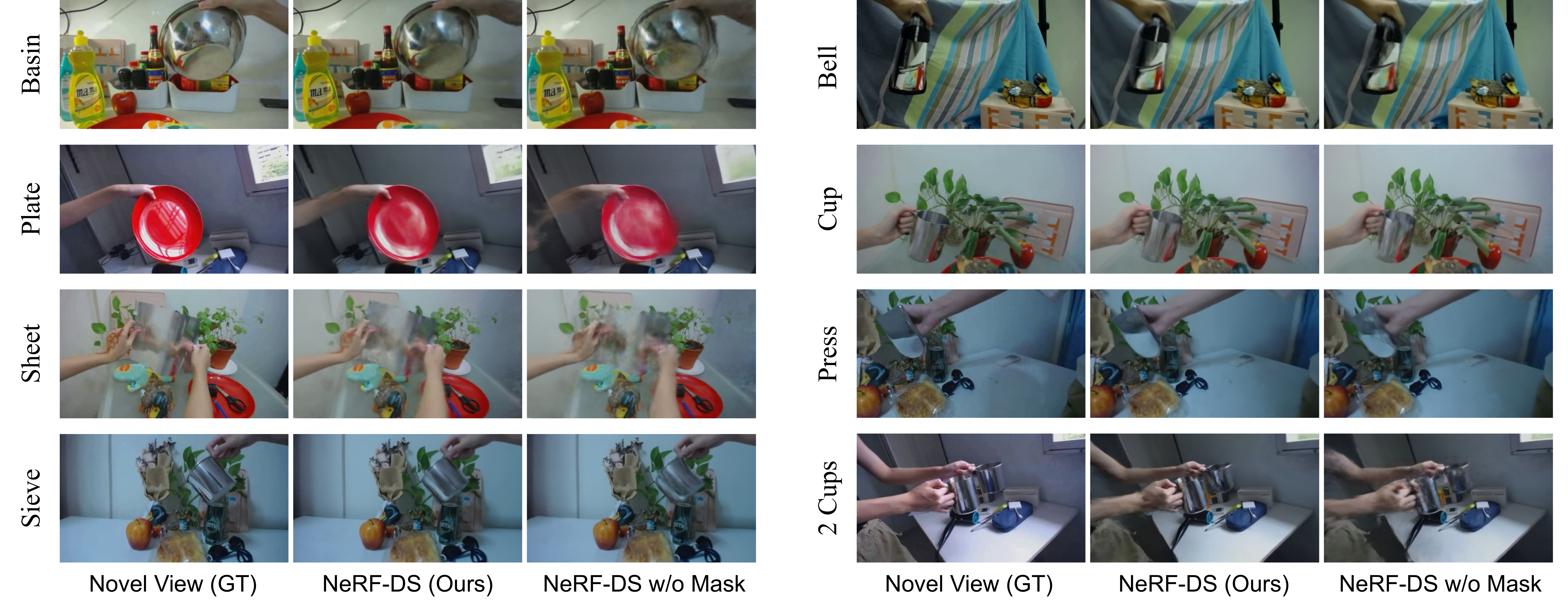}
   \caption{Qualitative comparison between our full model (NeRF-DS) and the ablation version without the surface-aware dynamic NeRF (NeRF-DS w/o Mask). }
   \label{fig:qualitative_mso}
\end{figure*}

\begin{figure*}[h]
  \centering
   \includegraphics[width=1.0\textwidth]{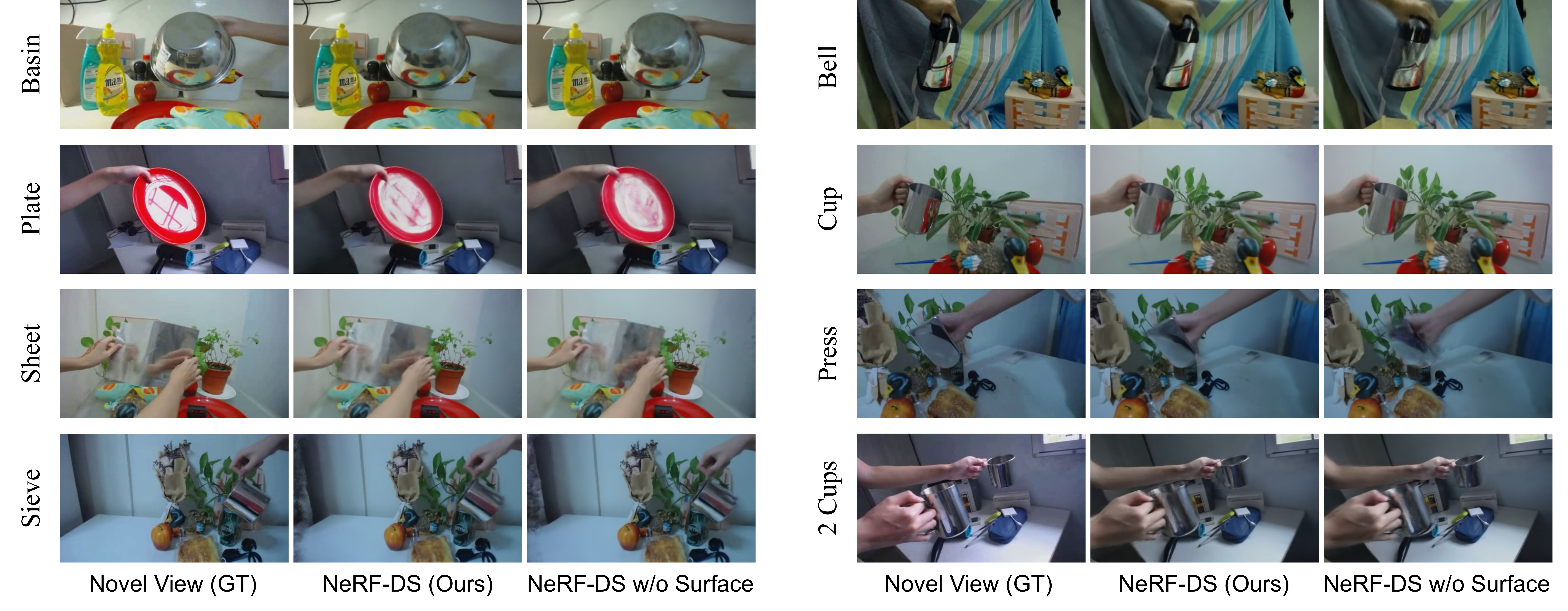}
   \caption{Qualitative comparison between our full model (NeRF-DS) and the ablation version without the mask guided deformation field (NeRF-DS w/o Surface). }
   \label{fig:qualitative_ref}
\end{figure*}

\begin{figure*}[th]
  \centering
   \includegraphics[width=0.9\textwidth]{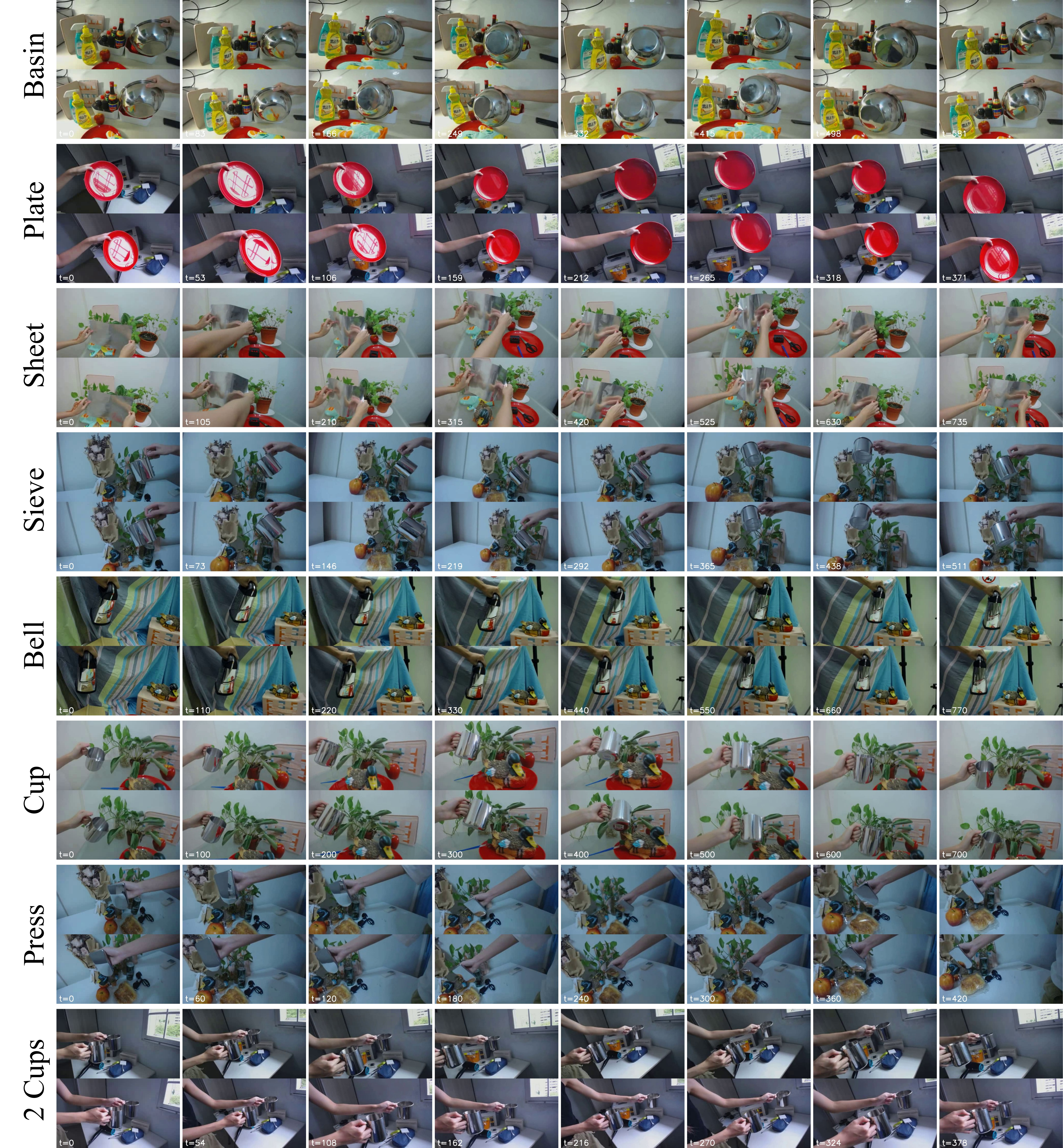}
   \caption{A snippet of the dynamic specular dataset for both cameras in 8 scenes. The training camera video is shown on the top and the test camera video is shown on the bottom.}
   \label{fig:dataset}
\end{figure*}

\section{Ablation Qualitative Results}

We present the qualitative comparison between the full and ablation versions of our models. The comparison between our NeRF-DS model and the ablation version without surface-aware dynamic NeRF is shown in~\cref{fig:qualitative_mso}. The comparison between our NeRF-DS model and the ablation version without mask guided deformation field is shown in~\cref{fig:qualitative_ref}.

\section{Additional Experiment Results}
We use delayed positional encoding for the spatial location $\mathbf{x}$ and sharp volumetric weights $w_i'$ for the mask rendering. In this section, we present additional ablation experiments to determine the best hyper-parameters for these two techniques.

We evaluate the performance of the NeRF-DS model on the ``Sheet" scene in the dynamic specular dataset, under different annealing strategy of the positional encoding for the observation space spatial coordinate $\mathbf{x}$ before it is fed to the NeRF color branch. Specifically, we evaluate the reconstruction performance with different schedules for the annealing coefficient $\tau$ of the $j$th term in the position encoding as shown in:
\begin{equation}
\label{eq:anneal}
    z_j(\tau)=\frac{1-cos(\pi \cdot clamp(\tau-j,0,1))}{2}   .
\end{equation}
We present the quantitative results in~\cref{tab:x_alpha_ablation}. Supported by the quantitative results, we choose to delay the use of $\mathbf{x}$ in the NeRF color branch for the first 50k iterations, and slowly increase the bandwidth to a maximum of 4 during the next 50k iterations.

We also evaluate the performance of the NeRF-DS model on the ``Sheet" scene in the dynamic specular dataset, with different sharp weights $w_i'$ for mask rendering. Particularly, we evaluate the reconstruction performance with different schedules for decreasing standard deviation $\beta$ in the Gaussian filter $\mathcal{N}(k_{\text{max}},\beta)$ applied to weight $w_i$ based on its ray distance $k_i$:
\begin{equation}
\label{eq:sharp}
    w_i^* = w_i \cdot P(k_i|\mathcal{N}(k_{\text{max}},\beta)),  w_i'=w_i^*/(\sum_{j}w_j^*)  .
\end{equation}
We present the quantitative results in~\cref{tab:mask_sharp_ablation}. Supported by the quantitative results, we choose to gradually decrease standard deviation for sharp mask weights from 1 to 0.1 during the first 30k iterations of the training. 

Additionally, we evaluate the performance of the NeRF-DS model on the ``Sheet" scene in the dynamic specular dataset, with surface normal $\mathbf{n}$ calculated from different spaces. The surface normal in the observation space used in our main results are warped from the surface normal in the canonical space to ensure cross frame consistency, \ie ${\mathbf{n}=\mathrm{R}^\top \mathbf{n}'}$. We compare the results with the model using surface normal calculated in the canonical space and the surface normal directly calculated in observation space as shown in~\cref{tab:norm_ablation}. The canonical space normal means $\mathbf{n}=\mathbf{n'}$. The observation space normal means the normal is supervised by the gradient of density with respect to the spatial coordinate in observation space:
\begin{subequations}
\begin{eqnarray}
    &\hat{\mathbf{n}}=-\frac{\nabla\sigma(\mathbf{x})}{\Vert\nabla\sigma(\mathbf{x})\Vert},\\
    &\mathcal{L}_{norm}=\sum_{i}w_i\Vert \mathbf{n}-\hat{\mathbf{n}}\Vert^2.
\end{eqnarray}
\end{subequations}
Supported by the quantitative comparison, we choose to use the surface normal warped from the canonical space for the better consistency over time. 

To demonstrate that our model has comparable performance to the baselines on non-specular dynamic scenes, we also present the experiment results of our model in the released scenes in the HyperNeRF dataset in~\cref{tab:hypernerf_data}. The results shown for Nerfies\cite{park2021nerfies} and HyperNeRF\cite{park2021hypernerf} are taken from the original paper, while the performance of our model is reproduced on the same data. Please note that due to our limited hardware (compared to the 4 TPU used in the original paper), our model trained on this HyperNeRF\cite{park2021hypernerf} dataset is using 1/10 of the batch size and 10 times the number of iterations. The performance comparison in this way is slightly in our disadvantages, as our reproduced HyperNeRF\cite{park2021hypernerf} models under this setting perform worse than the reported models.

\begin{table}[]
\scriptsize
\begin{tabular}{|l|ccc|}
\hline
    Positional Encoding Annealing                & MS-SSIM$\uparrow$             & PSNR$\uparrow$               & LPIPS$\downarrow$             \\ \hline
constant 4                              & 0.911                         & 25.4                         & 0.118                         \\
increase to 4 for 50k iter.             & 0.915                         & \cellcolor[HTML]{FCE4D6}25.7 & 0.119                         \\
delay 10k, increase to 4 for 50k iter.  & 0.914                         & 25.6                         & 0.121                         \\
delay 50k, increase to 4 for 50k iter.  & \cellcolor[HTML]{FCE4D6}0.918 & \cellcolor[HTML]{FCE4D6}25.7 & \cellcolor[HTML]{FCE4D6}0.115 \\
delay 100k, increase to 4 for 50k iter. & \cellcolor[HTML]{FFF2CC}0.917 & \cellcolor[HTML]{FCE4D6}25.7 & \cellcolor[HTML]{FFF2CC}0.117 \\
without x                                    & 0.913                         & 25.5                         & 0.120                         \\ \hline
\end{tabular}
\caption{Quantitative results on different annealing strategy for adding observation space coordinate $\mathbf{x}$ to the color branch of the canonical NeRF. Experiments are performed on the ``Sheet" scene. The \colorbox{myred}{best} and \colorbox{myyellow}{second best} results are color coded.}
\label{tab:x_alpha_ablation}
\end{table}

\begin{table}[]
\footnotesize
\begin{tabular}{|l|ccc|}
\hline
Standard Deviation Schedule & MS-SSIM$\uparrow$             & PSNR$\uparrow$               & LPIPS$\downarrow$             \\ \hline
1 to 0.01 for 30k iter.     & 0.916                         & \cellcolor[HTML]{FCE4D6}25.8 & 0.122                         \\
1 to 0.03 for 30k iter.     & 0.905                         & 25.3                         & 0.125                         \\
1 to 0.1 for 30k iter.      & \cellcolor[HTML]{FCE4D6}0.918 & \cellcolor[HTML]{FFF2CC}25.7 & \cellcolor[HTML]{FCE4D6}0.115 \\
1 to 0.3 for 30k iter.      & \cellcolor[HTML]{FFF2CC}0.917 & \cellcolor[HTML]{FFF2CC}25.7 & \cellcolor[HTML]{FFF2CC}0.120 \\
without sharping                 & 0.909                         & 25.6                         & 0.126                         \\ \hline
\end{tabular}
\caption{Quantitative results on different schedule for decreasing the standard deviation $\beta$ for the Gaussian filter to sharp the mask weights. Experiments are performed on the ``Sheet" scene. The \colorbox{myred}{best} and \colorbox{myyellow}{second best} results are color coded.}
\label{tab:mask_sharp_ablation}
\end{table}

\begin{table}[]
\footnotesize
\begin{tabular}{|l|ccc|}
\hline
Surface Normal              & MS-SSIM$\uparrow$             & PSNR$\uparrow$               & LPIPS$\downarrow$             \\ \hline
Warped from canonical space & \cellcolor[HTML]{FCE4D6}0.918 & \cellcolor[HTML]{FCE4D6}25.7 & \cellcolor[HTML]{FCE4D6}0.115 \\
Canonical space normal      & \cellcolor[HTML]{FFF2CC}0.913 & 25.5                         & 0.119                         \\
Observation space normal    & \cellcolor[HTML]{FFF2CC}0.913 & \cellcolor[HTML]{FFF2CC}25.6 & \cellcolor[HTML]{FFF2CC}0.117 \\ \hline
\end{tabular}
\caption{Quantitative results on types of surface normal $\mathbf{n}$ used. Experiments are performed on the ``Sheet" scene. The \colorbox{myred}{best} and \colorbox{myyellow}{second best} results are color coded.}
\label{tab:norm_ablation}
\end{table}

\begin{table}[]
\centering
\scriptsize
\begin{tabular}{l|p{6mm}p{6mm}p{6mm}p{6mm}|p{6mm}}
\hline
               & Printer    & Broom         & Chicken       & Banana        & Mean       \\
               & \tiny{PSNR$\uparrow$}          & \tiny{PSNR$\uparrow$}          & \tiny{PSNR$\uparrow$}          & \tiny{PSNR$\uparrow$}        & \tiny{PSNR$\uparrow$}          \\ \hline
Nerfies~\cite{park2021nerfies}        & 20.0          & 19.3          & 26.9          & 23.3         & 22.4          \\
HyperNeRF~\cite{park2021hypernerf}      & 20.0          & \textbf{20.6} & 27.6 & \textbf{24.3}  & \textbf{23.1} \\ \hline
NeRF-DS (Ours) & \textbf{21.0} & 19.6          & \textbf{27.9}          & 22.8          & 22.8          \\ \hline
\end{tabular}
\vspace{-2mm}
\caption{Performance on non-specular HyperNeRF~\cite{park2021hypernerf} dataset.}
\vspace{-6mm}
\label{tab:hypernerf_data}
\end{table}

\section{Additional Qualitative Analysis}
\begin{figure}[]
  \centering
  \includegraphics[width=0.8\linewidth]{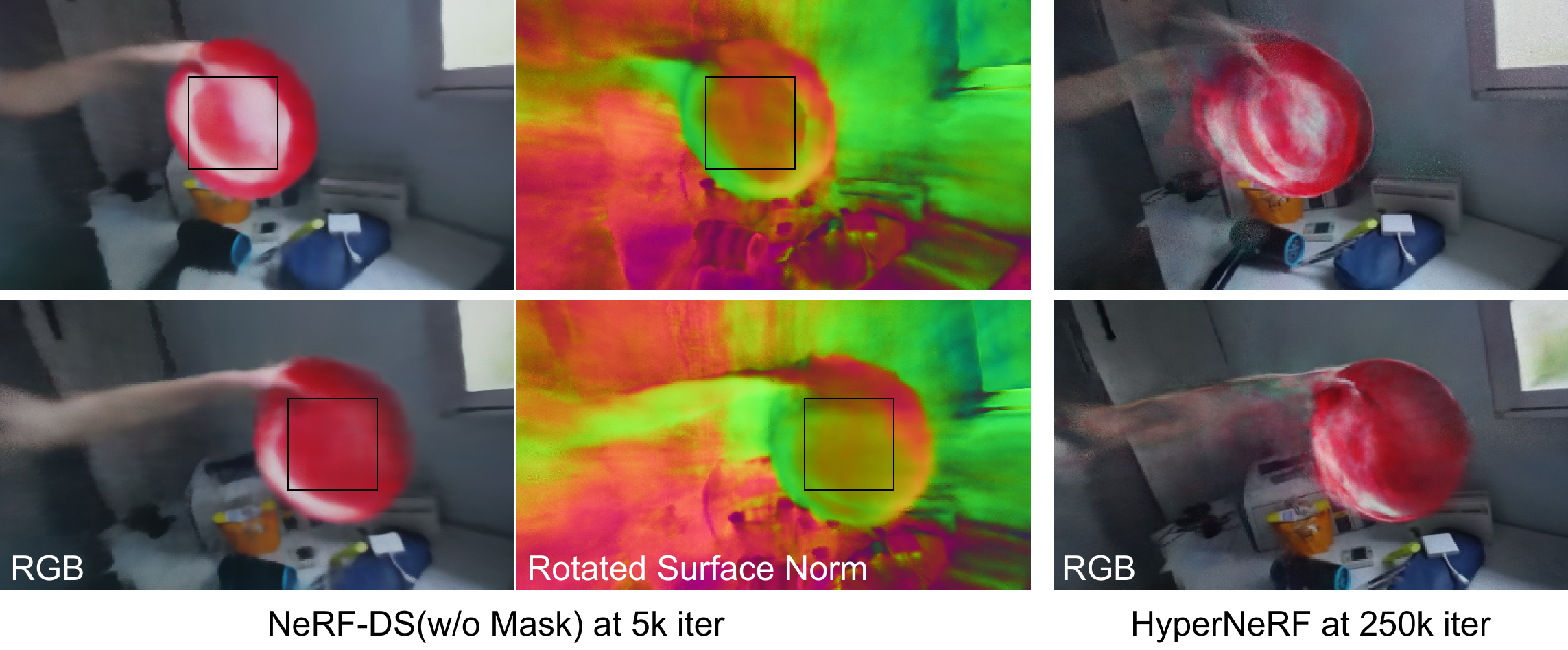}
  \vspace{-4mm}
  \caption{A case study on how different surface norms can guide rendering different reflected colors.}
  \label{fig:norm-case-study}
  \vspace{-3mm}
\end{figure}

To further analyse the influence of the surface normal input on the rendering, we present a qualitative case study. Taking the early stage result of NeRF-DS (w/o mask) as an example (\cref{fig:norm-case-study}), the norms predicted for the middle part of the plate in two frames are different. With this input, our NeRF-DS model can render different reflected colors of the same surface. However, HyperNeRF fails to recognize the surfaces in the two frames to be the same and renders severe geometric artifacts. Additional masks can further suppress the geometric artifacts, but our ablation study suggests that the surface normal alone also contributes significantly to the performance (20.3\% LPIPS improvement from baseline).

\section{Dynamic Specular Dataset Details}
The dataset consists of 8 scenes of various dynamic specular objects in everyday environments. Two rigidly connected cameras are used to capture the scenes for 480x270 resolution. Different types of objects and surfaces are used as shown in~\cref{tab:dataset}. A snippet of the dataset is shown in~\cref{fig:dataset}. We appreciate the help of Liu Shiru and CVRP lab members for collecting this dataset.

\begin{table}[]
\footnotesize
\begin{tabular}{|l|c|l|}
\hline
\multicolumn{1}{|c|}{Scene Name} & $\#$ frames & \multicolumn{1}{c|}{Object Attribute} \\ \hline
Basin                            & 668      & Curved+Flat, Metallic                 \\
Plate                            & 424      & Curved+Flat, Plastic, Colored         \\
Sheet                            & 846      & Soft, Metallic, Non-Rigid Deformation \\
Sieve                            & 584      & Curved, Metallic, Porous Bottom       \\
Bell                             & 881      & Slightly Curved, Metallic             \\
Cup                              & 807      & Curved+Flat, Metallic                 \\
Press                            & 487      & Flat, Metallic                        \\
2 Cups                           & 437      & Curved+Flat, Metallic, 2 Objects      \\ \hline
\end{tabular}
\caption{Details of each scene in the dynamic specular dataset. }
\label{tab:dataset}
\end{table}

\end{document}